\documentclass[letterpaper, 10 pt, conference]{ieeeconf}  

\IEEEoverridecommandlockouts                              
\usepackage[table,xcdraw]{xcolor}
\usepackage{scalerel}
\usepackage{tikz}
\usetikzlibrary{svg.path}
\usepackage{amsmath}
\usepackage{colortbl}

\definecolor{orcidlogocol}{HTML}{A6CE39}
\tikzset{
  orcidlogo/.pic={
    \fill[orcidlogocol] svg{M256,128c0,70.7-57.3,128-128,128C57.3,256,0,198.7,0,128C0,57.3,57.3,0,128,0C198.7,0,256,57.3,256,128z};
    \fill[white] svg{M86.3,186.2H70.9V79.1h15.4v48.4V186.2z}
                 svg{M108.9,79.1h41.6c39.6,0,57,28.3,57,53.6c0,27.5-21.5,53.6-56.8,53.6h-41.8V79.1z M124.3,172.4h24.5c34.9,0,42.9-26.5,42.9-39.7c0-21.5-13.7-39.7-43.7-39.7h-23.7V172.4z}
                 svg{M88.7,56.8c0,5.5-4.5,10.1-10.1,10.1c-5.6,0-10.1-4.6-10.1-10.1c0-5.6,4.5-10.1,10.1-10.1C84.2,46.7,88.7,51.3,88.7,56.8z};
  }
}

\newcommand\orcidicon[1]{\href{https://orcid.org/#1}{\mbox{\scalerel*{
\begin{tikzpicture}[yscale=-1,transform shape]
\pic{orcidlogo};
\end{tikzpicture}
}{|}}}}

\usepackage{hyperref}

\title{\LARGE \bf
  Improving Human Performance Using Mixed Granularity of Control\\
  in Multi-Human Multi-Robot Interaction} 
\author{Jayam Patel$^{1}$\orcidicon{0000-0002-0687-4169} and Carlo Pinciroli$^{1}$\orcidicon{0000-0002-2155-0445} 
\thanks{$^{1}$ Robotics Engineering, Worcester Polytechnic Institute, MA, USA. Email: {\sf \{jupatel,cpinciroli\}@wpi.edu}}%
}

\usepackage{graphicx}
\usepackage{subcaption}
\usepackage{multirow}
\usepackage{units}
\usepackage{hyperref}

\begin{document}

\maketitle
\thispagestyle{empty}
\pagestyle{empty}


\begin{abstract}
  Due to the potentially large number of units involved, the interaction with a
  multi-robot system is likely to exceed the limits of the span of apprehension
  of any individual human operator. In previous work, we studied how this issue
  can be tackled by interacting with the robots in two modalities ---
  environment-oriented and robot-oriented. In this paper, we study how this
  concept can be applied to the case in which multiple human operators perform
  supervisory control on a multi-robot system. While the presence of extra
  operators suggests that more complex tasks could be accomplished, little
  research exists on how this could be achieved efficiently. In particular, one
  challenge arises --- the \emph{out-of-the-loop performance problem} caused by
  a lack of engagement in the task, awareness of its state, and trust in the
  system and in the other operators. Through a user study involving 28 human
  operators and 8 real robots, we study how the concept of mixed granularity in
  multi-human multi-robot interaction affects user engagement, awareness, and
  trust while balancing the workload between multiple operators.
\end{abstract}

\section{Introduction}
    
Multi-robot systems will assist humans in accomplishing complex
tasks~\cite{Brambilla2013}, including home maintenance~\cite{gates2007robot},
mining~\cite{rubio2012mining}, bridge inspection~\cite{oh2009bridge}, disaster
recovery~\cite{murphy2014disaster,hamins2015research}, and colonizing the cosmos~\cite{goldsmith1999book}.
In all these applications, robot autonomy is a necessary but not sufficient
condition. Along with autonomy, an equally important component of these systems
is the ability for humans to supervise and affect the behavior of the robots
over the duration of the mission~\cite{kolling2015human}.

By their very nature, multi-robot systems are complex systems composed of many
interacting parts. As such, these systems often exceed the span of apprehension
of any individual human, which is typically limited to 7($\pm 2$)
entities~\cite{miller1956magical,lewis2010choosing}. This limits the performance
of the operators, which is typically measured in terms of workload, situational
awareness, and trust in the
system~\cite{endsley2017here,hussein2018mixed,chen2014human}. A natural approach
to improving human performance is to relieve the burden of individual operators
by conceiving of supervisory control schemes in which multiple operators
cooperate.

However, with multiple humans in the system, additional challenges arise, such
as coping with ineffective group dynamics~\cite{allen2004exploring}, unbalanced
workload~\cite{mcbride2011understanding,chen2014human}, and inhomogeneous
awareness~\cite{riley2005situation,lee2008review,parasuraman1997humans}. These
challenges coalesce in a common, undesirable phenomenon: the
\emph{out-of-the-loop (OOTL) performance problem}, caused by a lack of
engagement in the task, awareness of its state, and trust in the system and
other operators~\cite{endsley1995out,gouraud2017autopilot}.

Little research exists on these topics in the context of multi-robot systems. In
this paper, we extend our previous work on mixed-granularity control of
multi-robot systems to study how our approach affects the performance of
multiple human operators. In previous work, we showed that allowing an operator
to control two levels of granularity---the task goal (by modifying the
environment) and the individual robots---improves the performance of the human
operator~\cite{patel2019}. In this study, we explore how this modality of
interaction affects the performance of multiple cooperating operators.

This paper offers two main contributions:
\begin{itemize}
\item From the technological point of view, we created the first
  mixed-granularity interface for multi-human-multi-robot interaction. Our
  interface is based on a networked augmented reality application that allows
  the operators to visualize and modify the global state of the system
  collaboratively on common tablets and smartphones.
\item From the scientific point of view, we assessed the validity of our
  approach through a user study of our interface in terms of workload, trust,
  and task performance. The user study involved 14 teams of 2 operators each,
  controlling a team of 8 robots in a collective construction scenario.
\end{itemize}
The paper is organized as follows. In Sec.~\ref{sec:background} we discuss
related work on human-robot interfaces.
In Sec.~\ref{sec:framework} we present our system and its design. In
Sec.~\ref{sec:userstudy} we detail our user study, followed by a discussion of
the results in Sec.~\ref{sec:discussion}. We conclude the paper in
Sec.~\ref{sec:conclusion}.

\section{Background}
\label{sec:background}
According to Endsley~\cite{endsley2017here}, granularity of control is a key
aspect affecting the OOTL performance problem. Low-level control includes robot selection
and manipulation~\cite{lewis_effects_2011,you2016curiosity,
  lewis_teams_2010,gromov_wearable_2016,kapellmann-zafra_human-robot_2016,alonso-mora_gesture_2015,nagi_human-swarm_2014,natraj_gesturing_2014,
  nagavalli2017multi}, while high-level control comprises of global goal
manipulation by defining navigation
goals~\cite{bashyal_human_2008,diaz-mercado_distributed_2015,kolling_human_2013,ayanian_controlling_2014},
team organization~\cite{johnson2008human,dias_dynamically_2006}, or allocating
tasks~\cite{malvankar-mehta_optimal_2015}. Limiting control to one type of
granularity creates a fundamental tradeoff~\cite{endsley2017here}.  Low-level
control offers more opportunity for interaction and sense of trust in the
system, but it causes higher workload and stress. Conversely, high-level control
limits the amount of workload, often leading to boredom and lower situational
awareness, which in turn results in the OOTL performance problem.

There exists little research on supervisory control in multi-human-multi-robot
systems. Several papers study the case in which humans play the role of a
bystanders in, e.g., navigation of robots in a shared
environment~\cite{huang_human-oriented_2010,claes_multi_2018,wang_trust-based_2018,tseng_multi-human_2014,higuera_socially-driven_2012,bajcsy_scalable_2018}
and human
tracking~\cite{chen_tun_chou_multi-robot_2011,ong_sensor_2012,ong_sensor_2012,zhang_optimal_2016,weinberg_creation_2009,iqbal_coordination_2017}. Other
works focus on how to coordinate teams of humans and
robots~\cite{beer_framework_2017, malik_developing_2019, you_teaming_2017,
  tsarouchi_humanrobot_2017, freedy_multiagent_2008, jones_dynamically_2006,
  johnson2008human}. In supervisory control, past work focused on investigating
the influence of autonomy and resource sharing on the task
performance~\cite{lewis_effects_2011,lewis_teams_2010}. Researchers also
investigated the effects of curiosity and training on increasing task
performance~\cite{you2016curiosity}. However, to the best of our knowledge,
there has not been any study on the out-of-the-loop performance problem in the
context of multi-robot systems.

\section{The Networked Augmented Reality App} \label{sec:framework}
\subsection{System Overview}
\begin{figure}[t]
  \centering
  \includegraphics[width=0.3\textwidth]{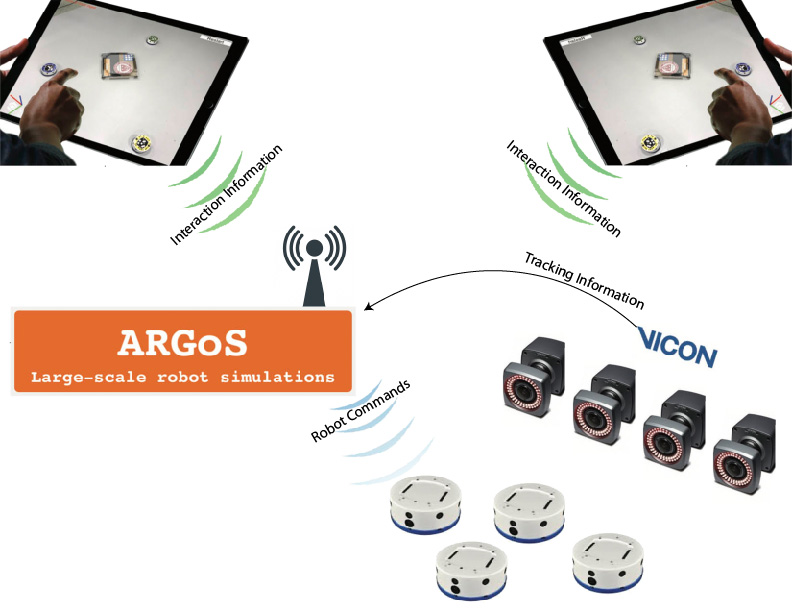}
  \caption{System overview.}
  \label{fig:system}
\end{figure}
The system is comprised of four components (see Fig.~\ref{fig:system}):
\begin{enumerate}
\item A distributed augmented reality interface implemented as an app for a
  hand-held device;
\item A team of robots pre-programmed with several autonomous behaviors,
  including a basic ``go-to-location'' and a more advanced ``collective
  transport''.
\item A Vicon motion tracking system for localizing the robots and dynamic
  objects in the environment;
\item ARGoS~\cite{Pinciroli:SI2012}, a multi-robot simulator acting as a
  middleware responsible for channeling data to the robots.
\end{enumerate}
The process starts when an operator
specifies a new position for an object, the selected team
of robots, or an individual robot on a hand-held device.
The hand-held device then broadcasts this information over the Wi-Fi
network for other active AR app users and ARGoS. The other AR apps process and
display the broadcasted change in the local augmented view. ARGoS generates and
broadcasts the goals for the robots, which execute the requested operations.\footnote{A video demonstration of the system is available at \\ \url{https://youtu.be/QGUYfBB9Ves}.}
    
\subsection{User Interface}
\begin{figure}[t]
  \centering
  \includegraphics[width=0.3\textwidth]{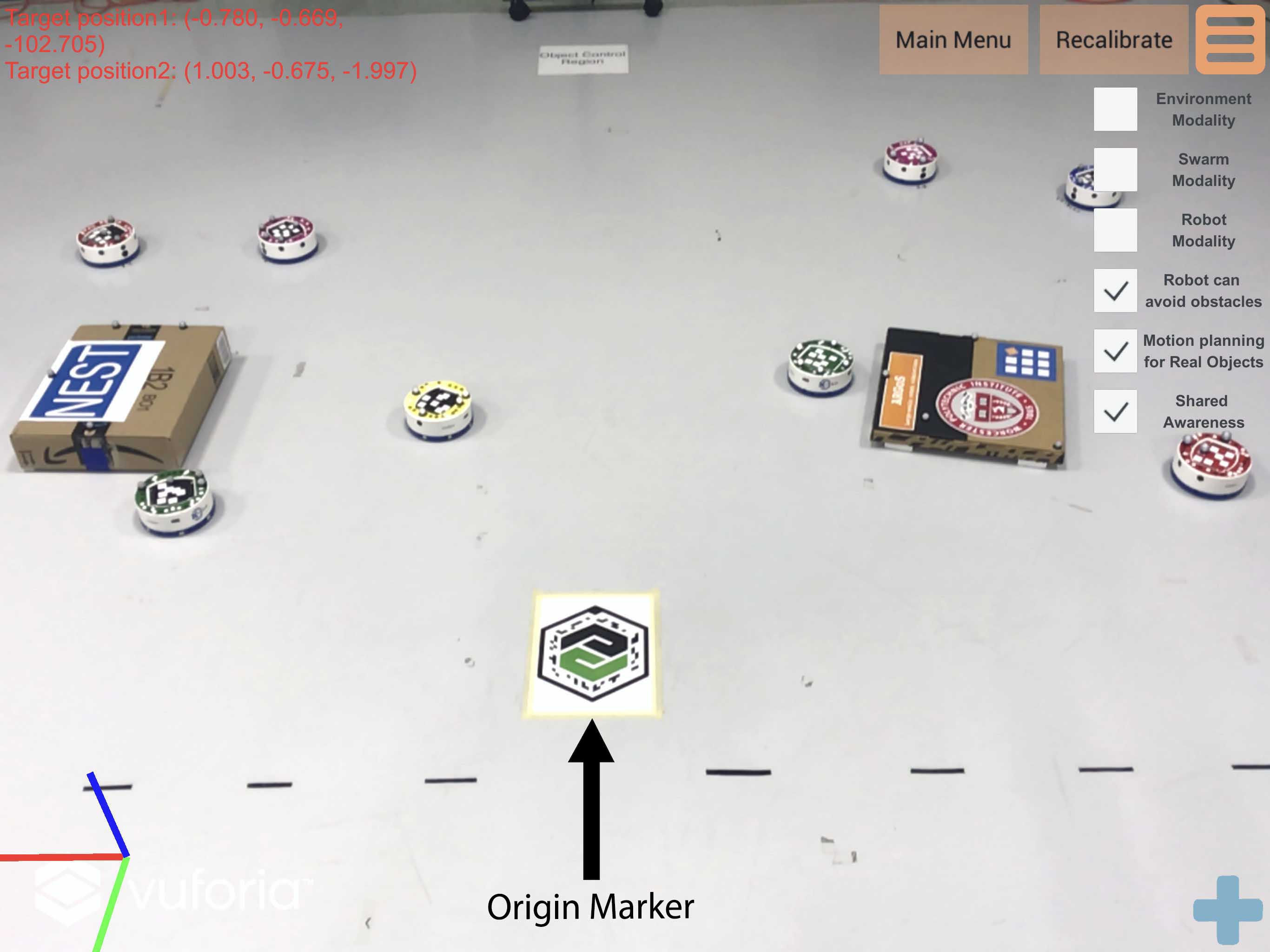}
  \caption{Screenshot of the AR Interface running on an iPad. The overlaid black
    arrow indicates the origin marker for initializing the coordinate frame of
    the interface.}%
  \label{fig:app}
\end{figure}
The interaction between operators and robots occurs through an Augmented Reality
(AR) application installed on handheld devices, such as smartphones or tablets. 
The app integrates Vuforia~\cite{vuforia}, a software development kit
for AR applications, and the Unity Game Engine~\cite{engine2008unity}. The
application can recognize the objects and the robots in real time using fiducial
markers. The operator can visualize and manipulate the identified objects and
robots by means of a virtual object overlayed on the real robot in the device
screen. The virtual object can be translated using a one-finger swipe and
rotated using a two-finger twist. The application also lets the operator select
some or all of the robots with a one-finger swipe. Fig.~\ref{fig:app} shows
the screenshot of the AR application. The top-left corner displays the desired
object position. The bottom-left corner depicts the current reference frame
based on the location of the operator and the unique origin marker. The
top-right corner offers the menu buttons for controlling additional
functionality such as re-calibration, toggling obstacle avoidance and toggling
visibility and detection of objects and robots. The bottom-right corner houses
the button for creating virtual objects dynamically.
    
\subsection{Interaction Modes}

\begin{figure}[t]
  \centering
  \begin{subfigure}[t]{0.2\textwidth}
    \includegraphics[width=\textwidth]{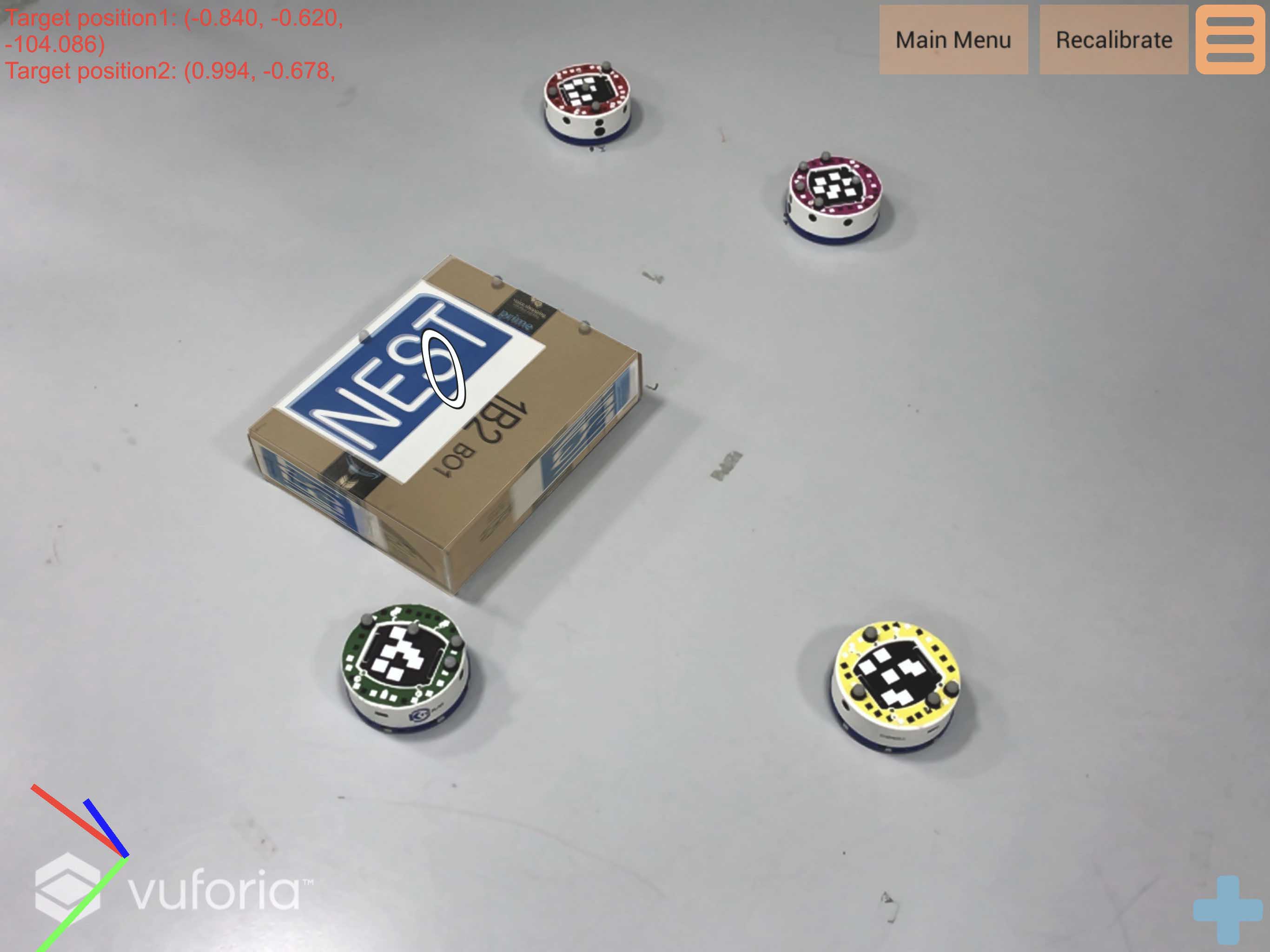}
    \caption{Object recognition}
    \label{fig:modeO1}
  \end{subfigure}
  \begin{subfigure}[t]{0.2\textwidth}
    \includegraphics[width=\textwidth]{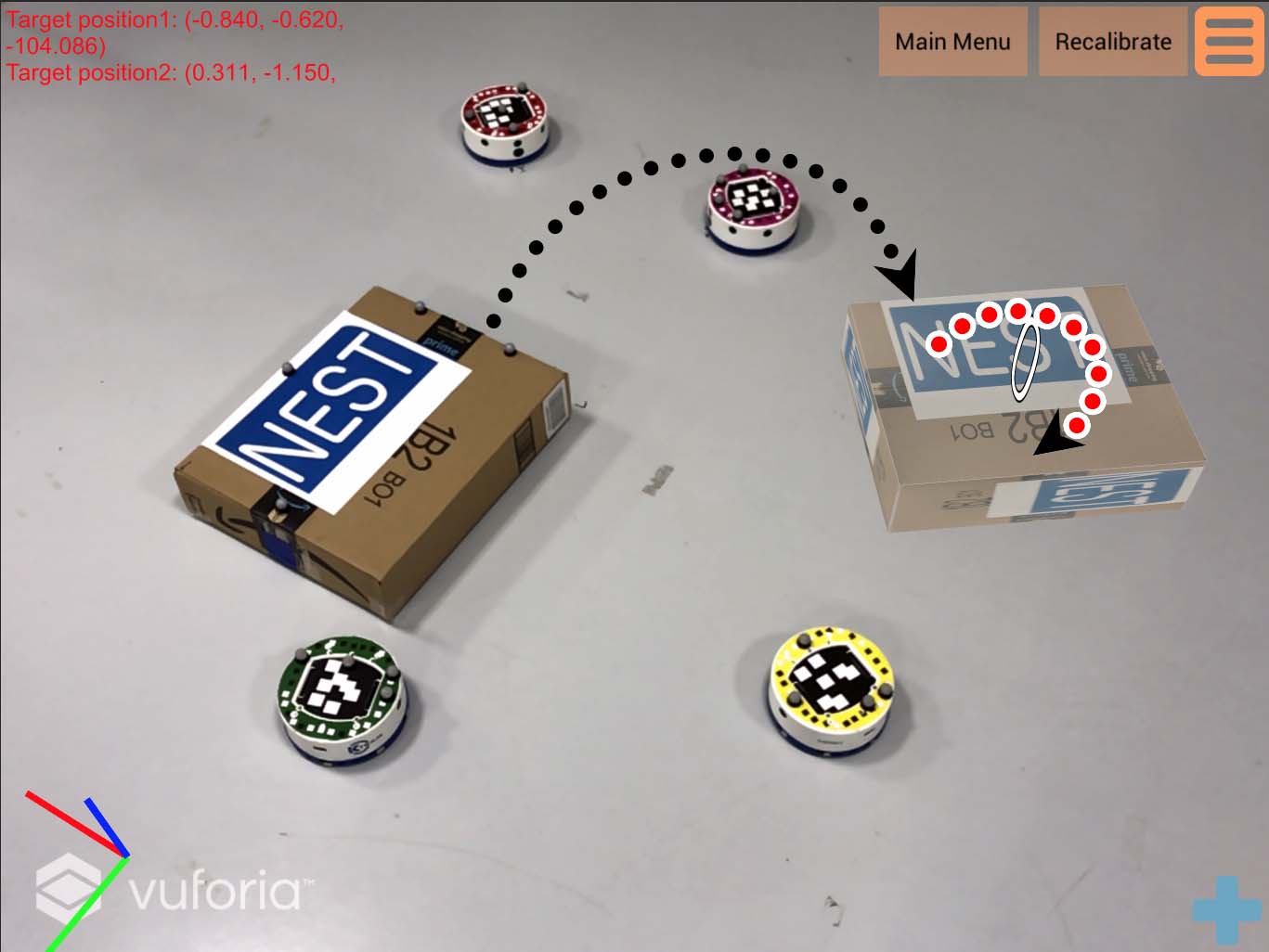}
    \caption{New Goal Defined}
    \label{fig:modeO2}
  \end{subfigure}
  \begin{subfigure}[t]{0.2\textwidth}
    \includegraphics[width=\textwidth]{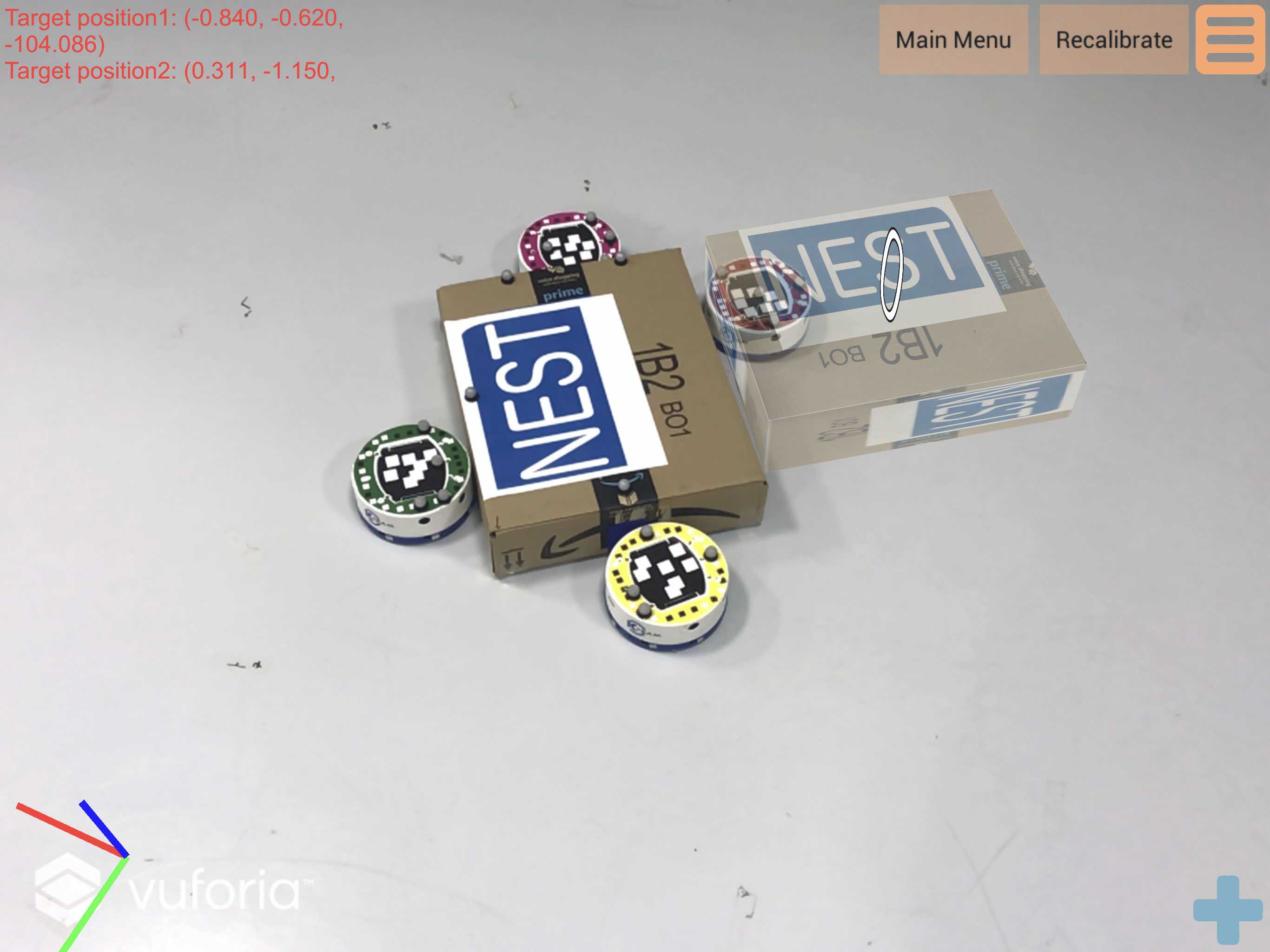}
    \caption{Robots approach and push}
    \label{fig:modeO3}
  \end{subfigure}
  \begin{subfigure}[t]{0.2\textwidth}
    \includegraphics[width=\textwidth]{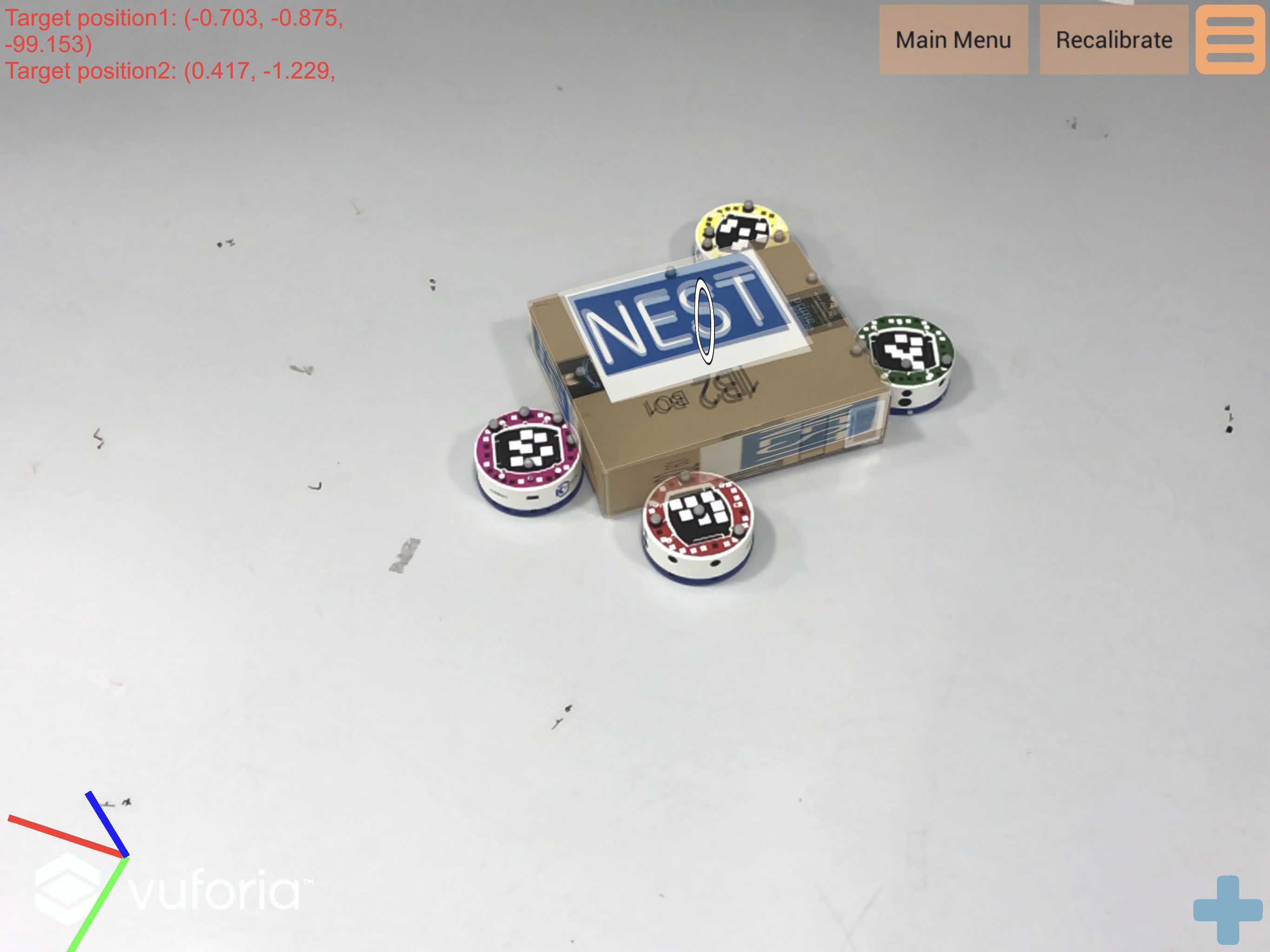}
    \caption{Transport complete}
    \label{fig:modeO4}
  \end{subfigure}
  \caption{Object manipulation by interaction with the virtual object through
    the interface. The overlaid dotted black arrow indicates the one-finger
    swipe gesture used to move the virtual object and the overlaid red dotted
     arrow indicates the two-finger rotation gesture.}\label{fig:modeO}
\end{figure}
            
\textbf{Object Manipulation.} The app overlays a virtual object over the real
object. The operator can manipulate this virtual object to define its desired
position. The app allows the operator to move multiple objects through this
gesture, and the respective team of robots will transport the objects in
parallel when enough robots are available. If the available robots are not
sufficient for a task, the app queues the request awaiting the completion of
prior tasks. If two or more operators simultaneously control the same object,
the app processes the latest broadcasted goal. In the current version of our
app, an operator must resolve this kind of conflicts through verbal
communication with other operators. Fig.~\ref{fig:modeO1} shows a virtual object
overlaid on the physical object. Fig.~\ref{fig:modeO2} demonstrates the rotation
and translation of the virtual object. Fig.~\ref{fig:modeO3} and
Fig.~\ref{fig:modeO4} show the robots performing the task.
        
\begin{figure}[t]
  \centering
  \begin{subfigure}[t]{0.2\textwidth}
    \includegraphics[width=\textwidth]{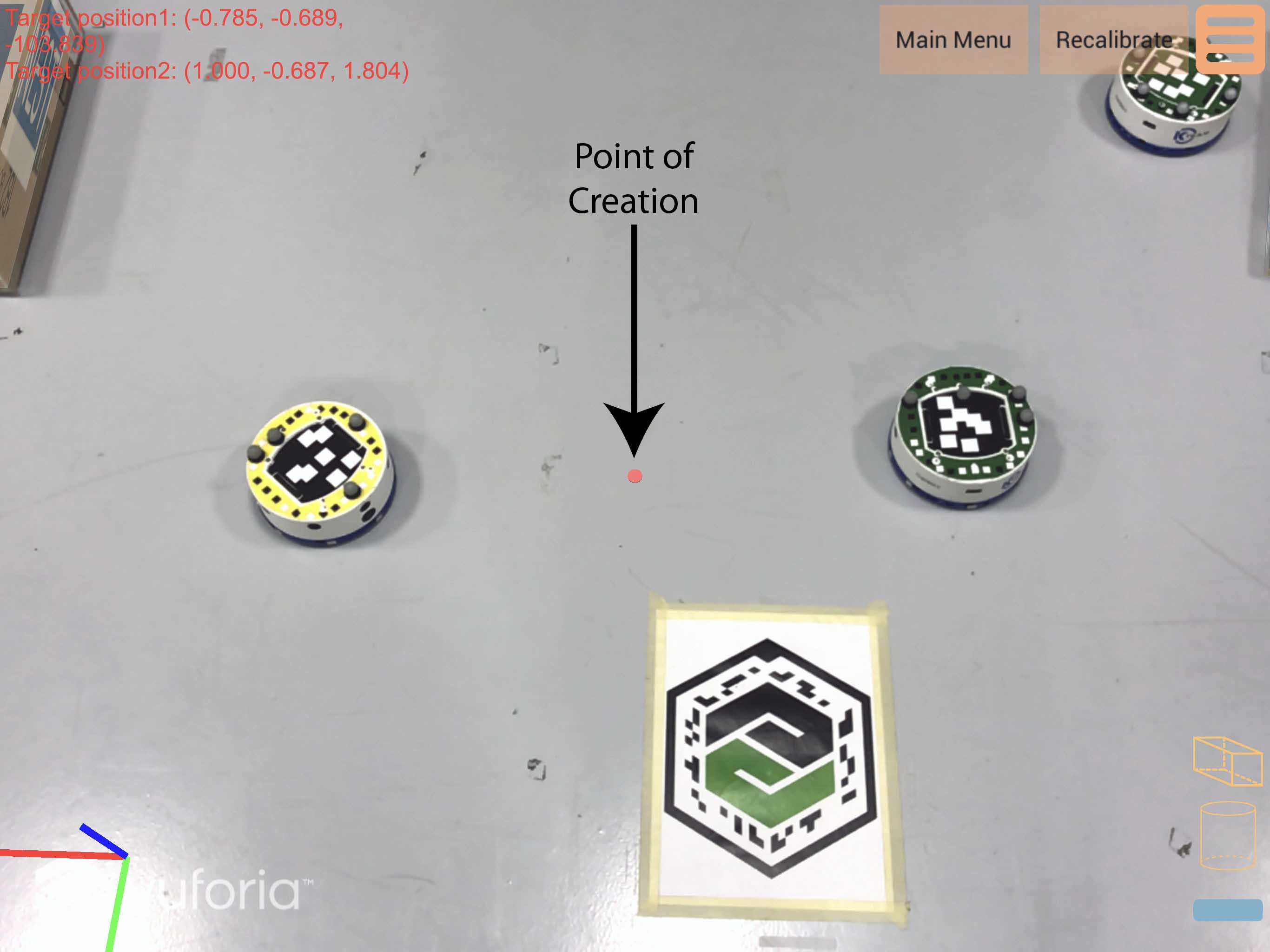}
    \caption{Virtual object creation mode}
    \label{fig:modeV1}
  \end{subfigure}
  \begin{subfigure}[t]{0.2\textwidth}
    \includegraphics[width=\textwidth]{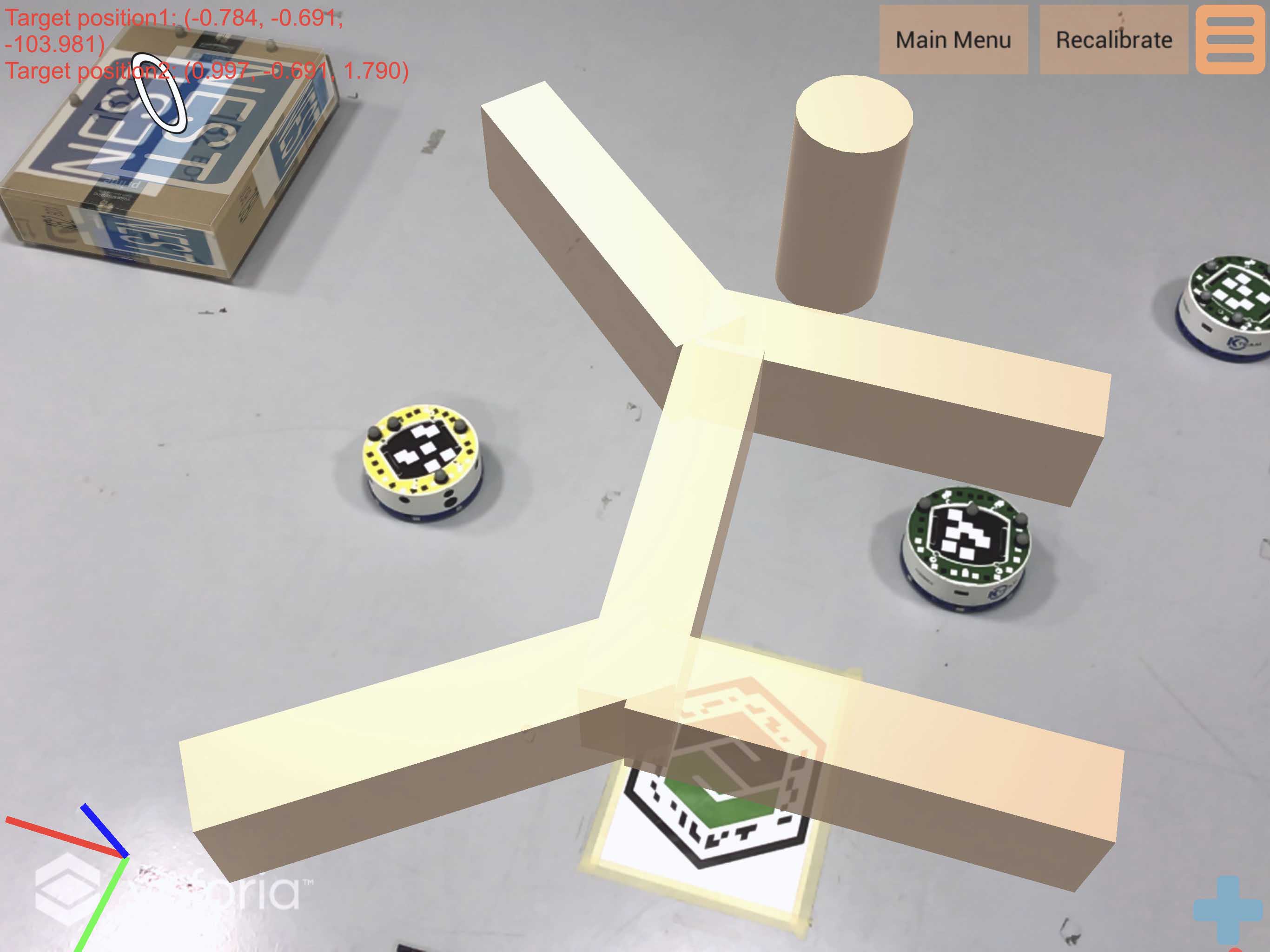}
    \caption{New virtual objects created and moved}
    \label{fig:modeV2}
  \end{subfigure}
  \caption{Virtual object creation and manipulation. The overlaid black arrow
    indicates the point of virtual object creation.}\label{fig:modeV}
\end{figure} 
        
\textbf{Virtual Object Creation, Manipulation and Deletion.} The app allows an operator to create virtual cuboids and virtual cylinders dynamically. The operator can reposition and reorient these objects. During virtual object creation, the app shows a point on the ground to signify the creation of virtual objects on that point (see Fig.~\ref{fig:modeV1}). The operator can delete the created virtual object with a two-finger long-press gesture. This modality is useful for creating virtual obstacles/walls and for defining a separate operating region for multiple operator scenario. Fig.~\ref{fig:modeV} shows the virtual objects arranged in the environment.
        
\begin{figure}[t]
  \centering
  \begin{subfigure}[t]{0.2\textwidth}
    \includegraphics[width=\textwidth]{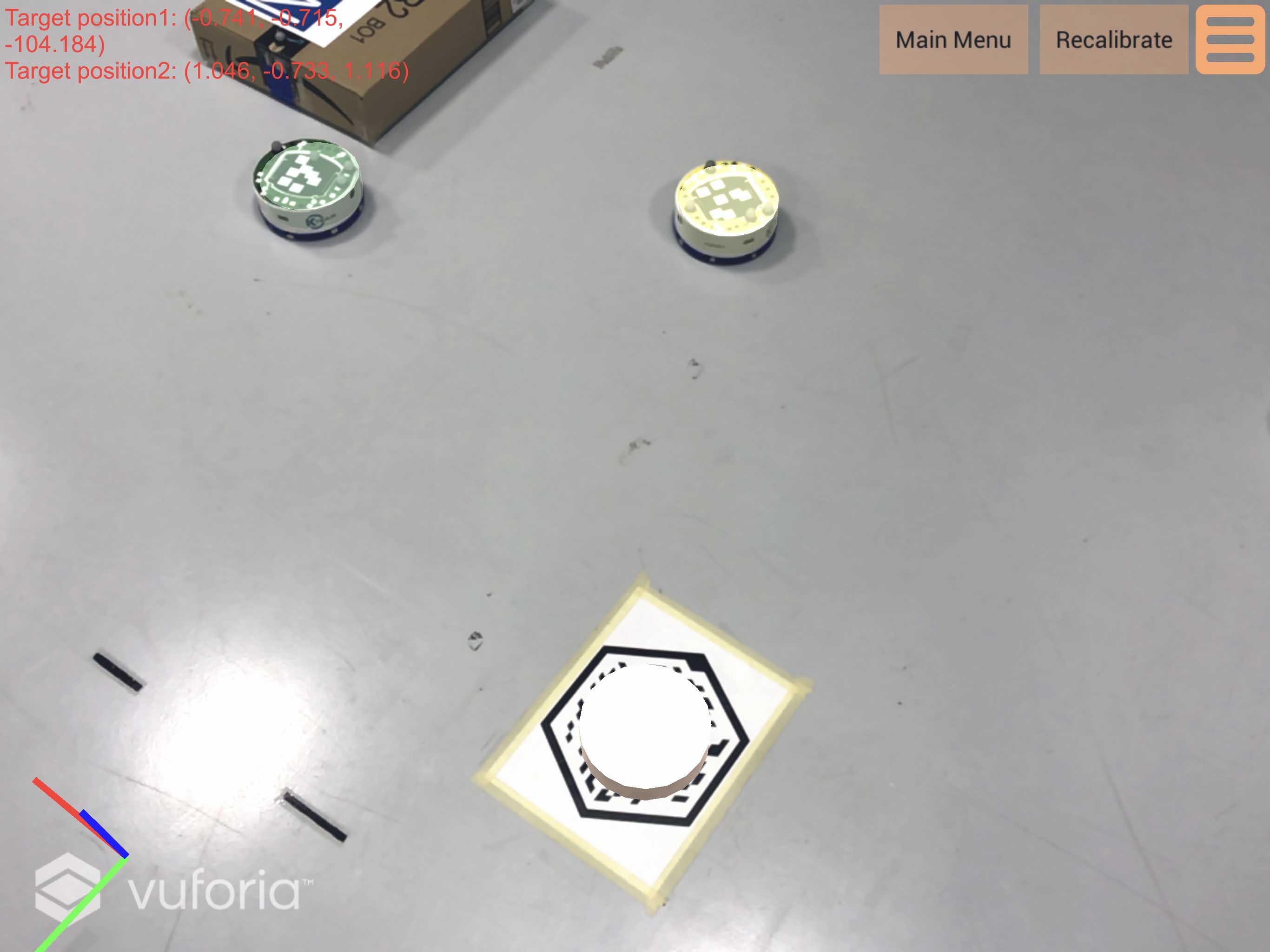}
    \caption{Robot recognition}
    \label{fig:modeR1}
  \end{subfigure}
  \begin{subfigure}[t]{0.2\textwidth}
    \includegraphics[width=\textwidth]{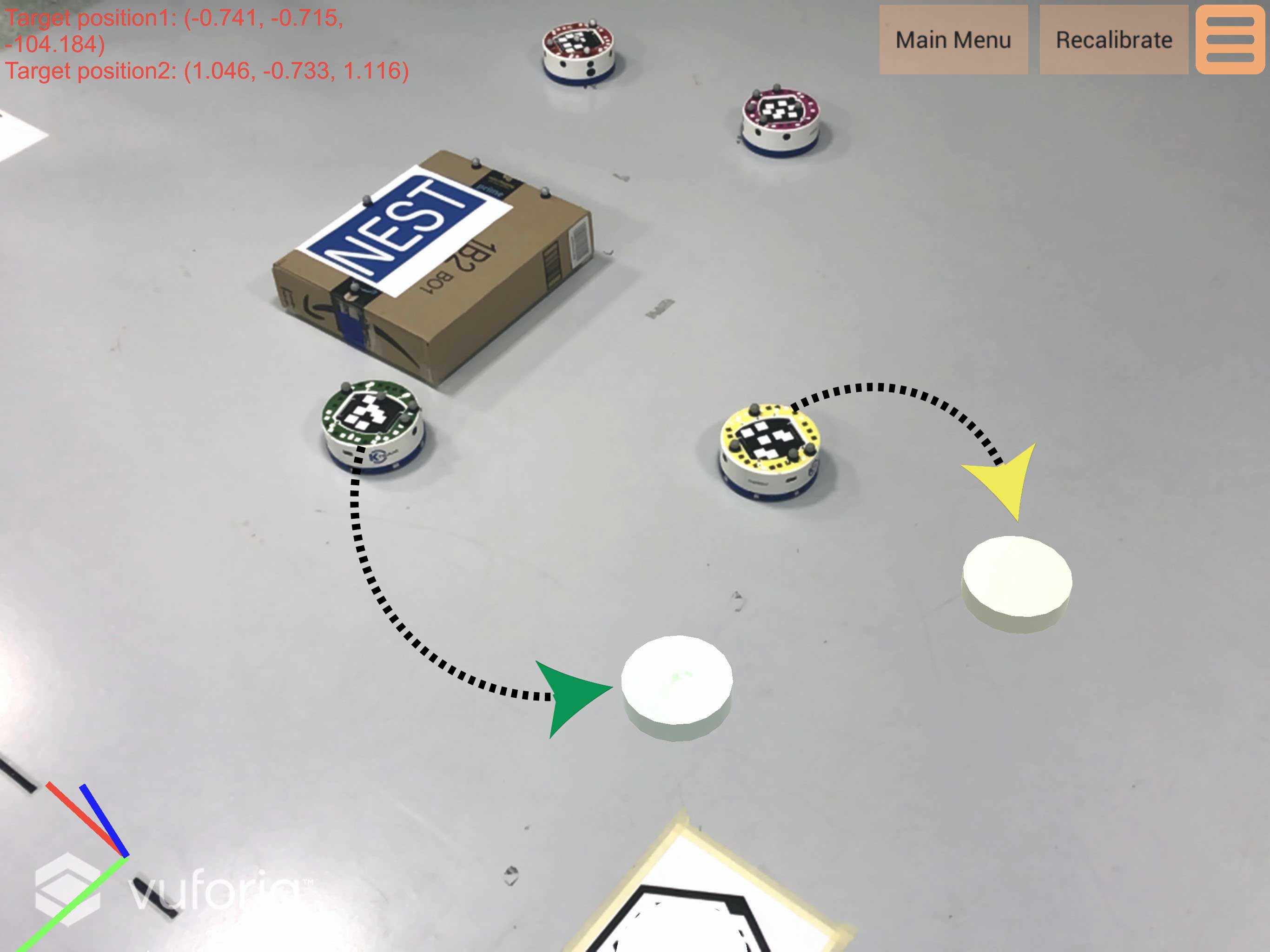}
    \caption{New robot position}
    \label{fig:modeR2}
  \end{subfigure}
  \caption{Robot manipulation by interacting with the virtual robots through the
    interface. The overlaid dotted black arrow indicates the one-finger swipe
    gesture to move the virtual robot and the arrowhead color indicates the
    moved virtual robots.}\label{fig:modeR}
\end{figure} 
    
\textbf{Robot Manipulation.} The app overlays a virtual robot over the real
robot. The operator can manipulate this virtual robot to define its desired
position. The color of the virtual robot resembles the color of the fiducial
markers to identify and differentiate between multiple robots. The app allows
the operator to move multiple robots through this gesture. Other robots
belonging to the same team pause their operation until the selected robot
achieves its defined position. If the robot is not part of any team, then the
position change does not hinder any other operation of the system. If two or
more operators simultaneously want to control the same robot, then the app
processes the newest broadcasted goal. Also in this case the operators can
resolve this kind of conflict through verbal
communication. Fig.~\ref{fig:modeR1} shows a virtual robot overlaid on the
physical robot. Fig.~\ref{fig:modeR2} illustrates the translation of the virtual
robot.
        
\begin{figure}[t]
  \centering
  \begin{subfigure}[t]{0.2\textwidth}
    \includegraphics[width=\textwidth]{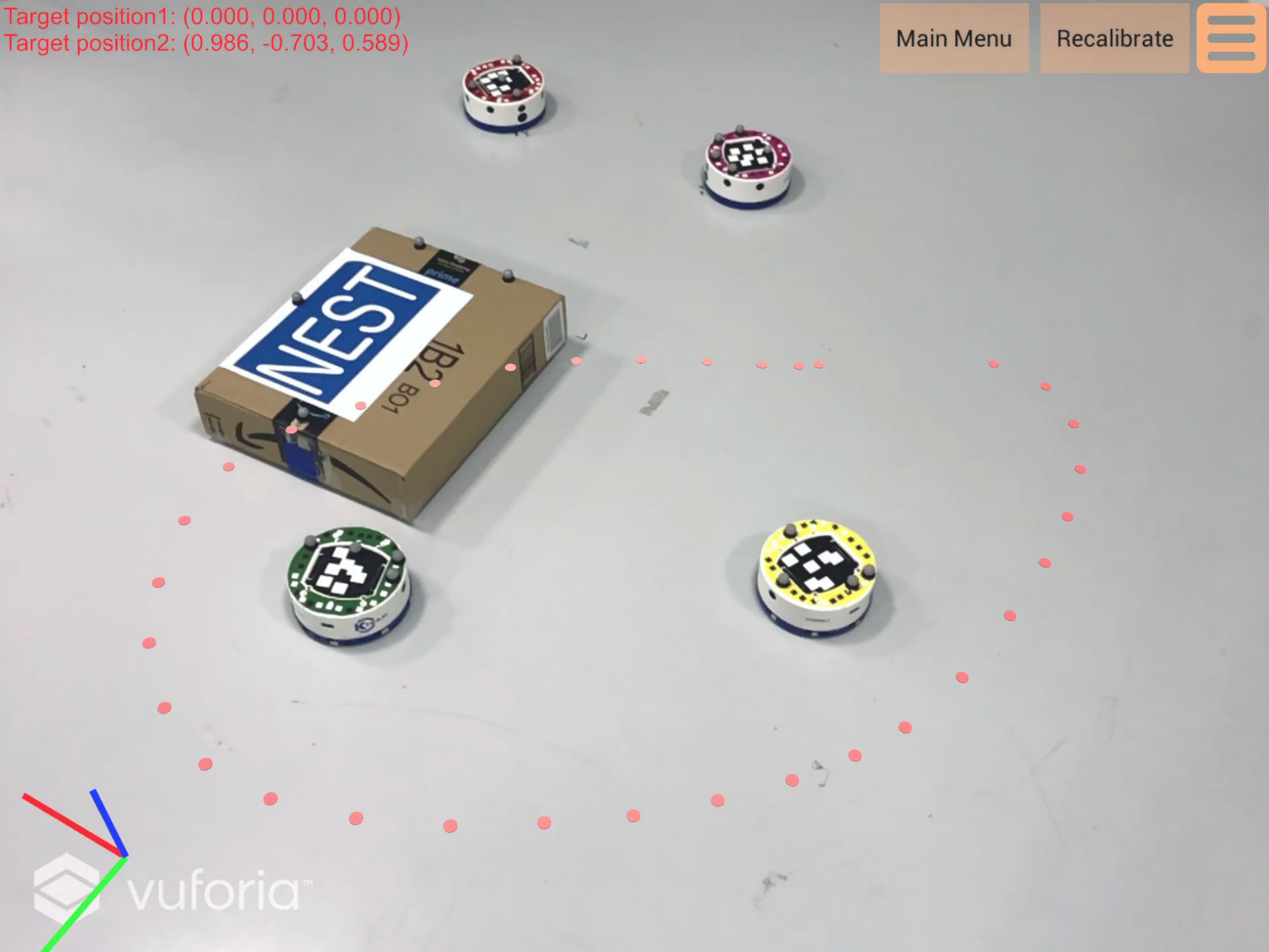}
    \caption{Robot team selection}
    \label{fig:modeS1}
  \end{subfigure}
  \begin{subfigure}[t]{0.2\textwidth}
    \includegraphics[width=\textwidth]{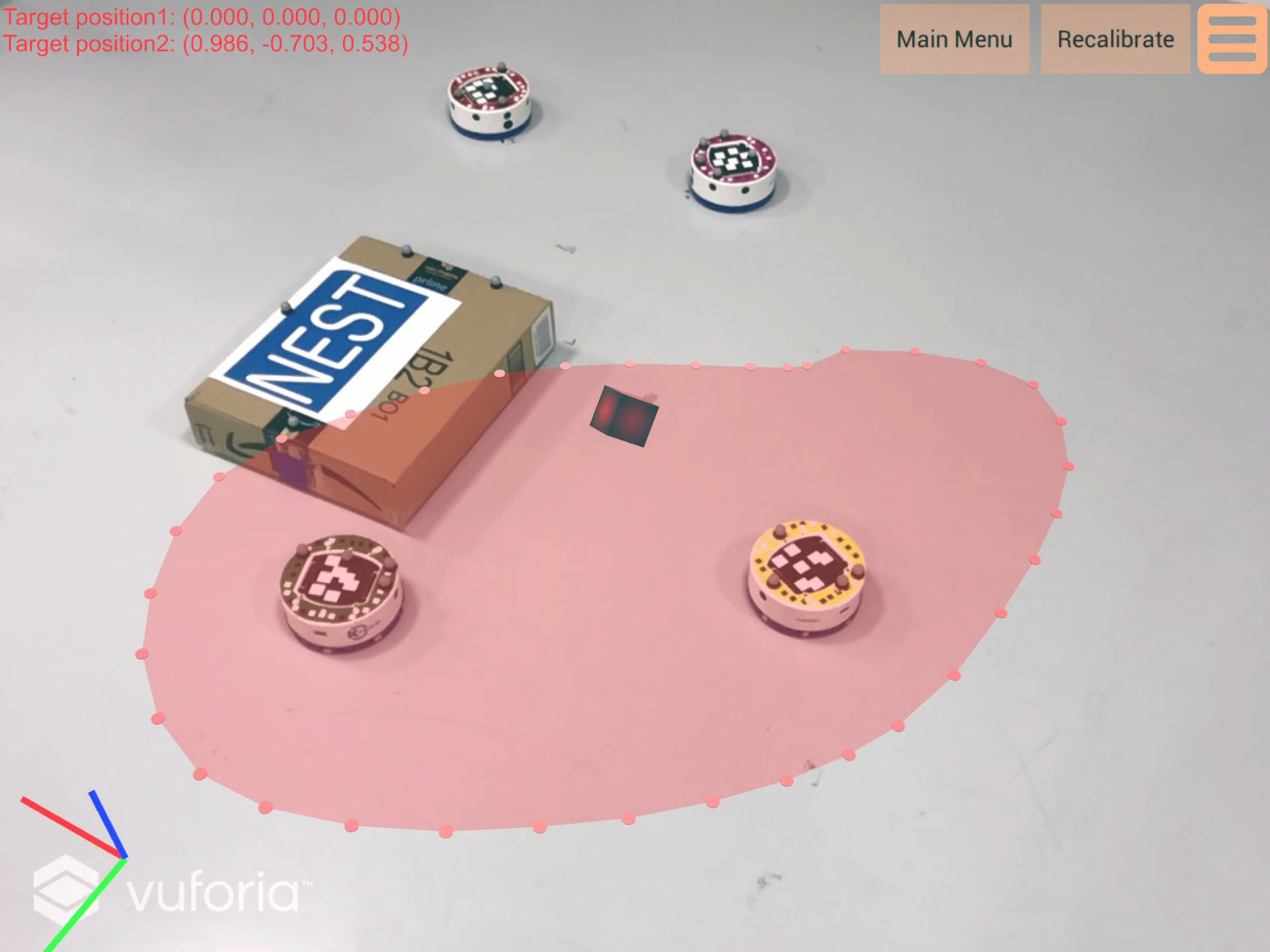}
    \caption{Robot team creation}
    \label{fig:modeS2}
  \end{subfigure}
  \begin{subfigure}[t]{0.2\textwidth}
    \includegraphics[width=\textwidth]{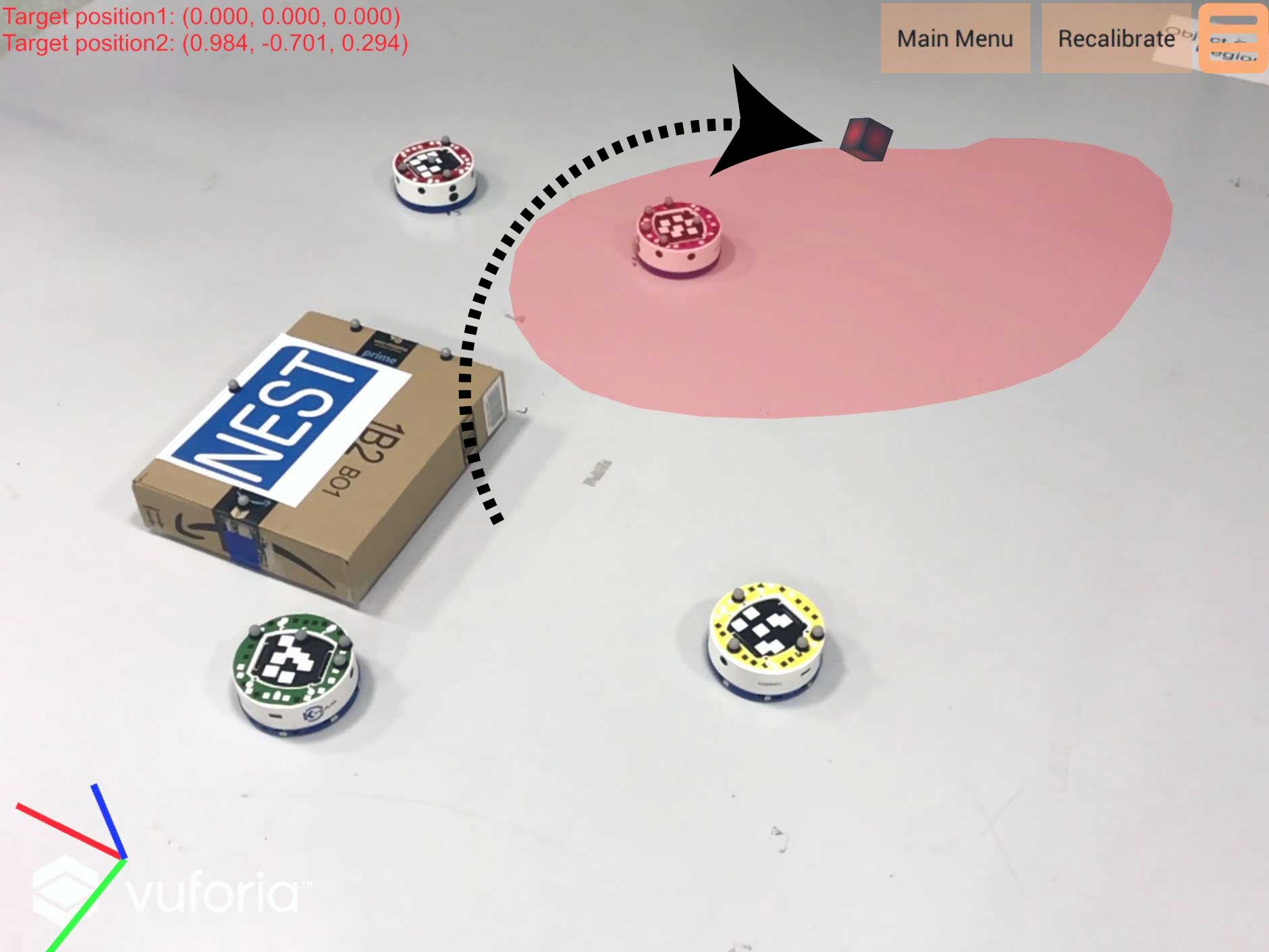}
    \caption{Robot team manipulation}
    \label{fig:modeS3}
  \end{subfigure}
  \begin{subfigure}[t]{0.2\textwidth}
    \includegraphics[width=\textwidth]{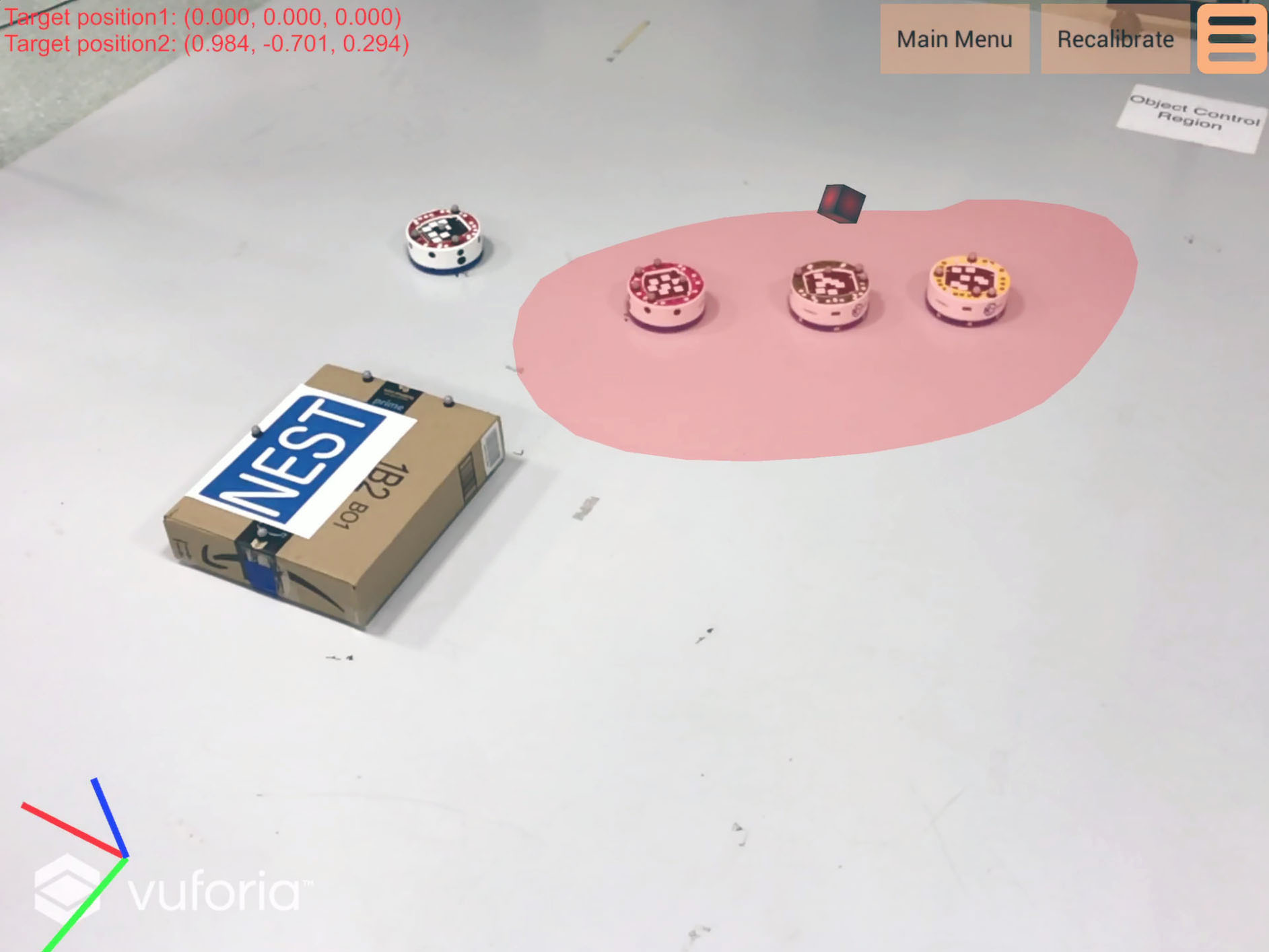}
    \caption{Robot team re-positioned}
    \label{fig:modeS4}
  \end{subfigure}
  \caption{Robot team creation and manipulation by interacting with the
    interface. The overlaid dotted black arrow indicates the one-finger swipe
    gesture to move the virtual cube for re-positioning the the team of
    robots.}\label{fig:modeS}
\end{figure} 
        
\textbf{Robot Team Selection and Manipulation.} With a one-finger swipe, the
operator can define an enclosed space for selecting all the robots physically
present in that region. A virtual layer overlays on the selected region and a
virtual cube appears at the centroid of this virtual layer. The operator can
manipulate this cube to define a goal position for all the robots in the
region. The robots then reposition themselves similarly to the Robot
Manipulation modality. An operator can select only one team of robots at a time,
and every time a new team of robots is selected, the app clears the last
selection. If two or more operators have the same robot in their selected team,
then the robot receives the most recent goal position. Fig.~\ref{fig:modeS1} and
Fig.~\ref{fig:modeS2} shows the selection of a group of
robots. Fig.~\ref{fig:modeS3} shows the manipulation of the virtual cube to
define a new goal position for the team of robots. Fig.~\ref{fig:modeS4}
shows the robots navigating to the desired position.
        
\begin{figure}[t]
  \centering
  \begin{subfigure}[t]{0.2\textwidth}
    \includegraphics[width=\textwidth]{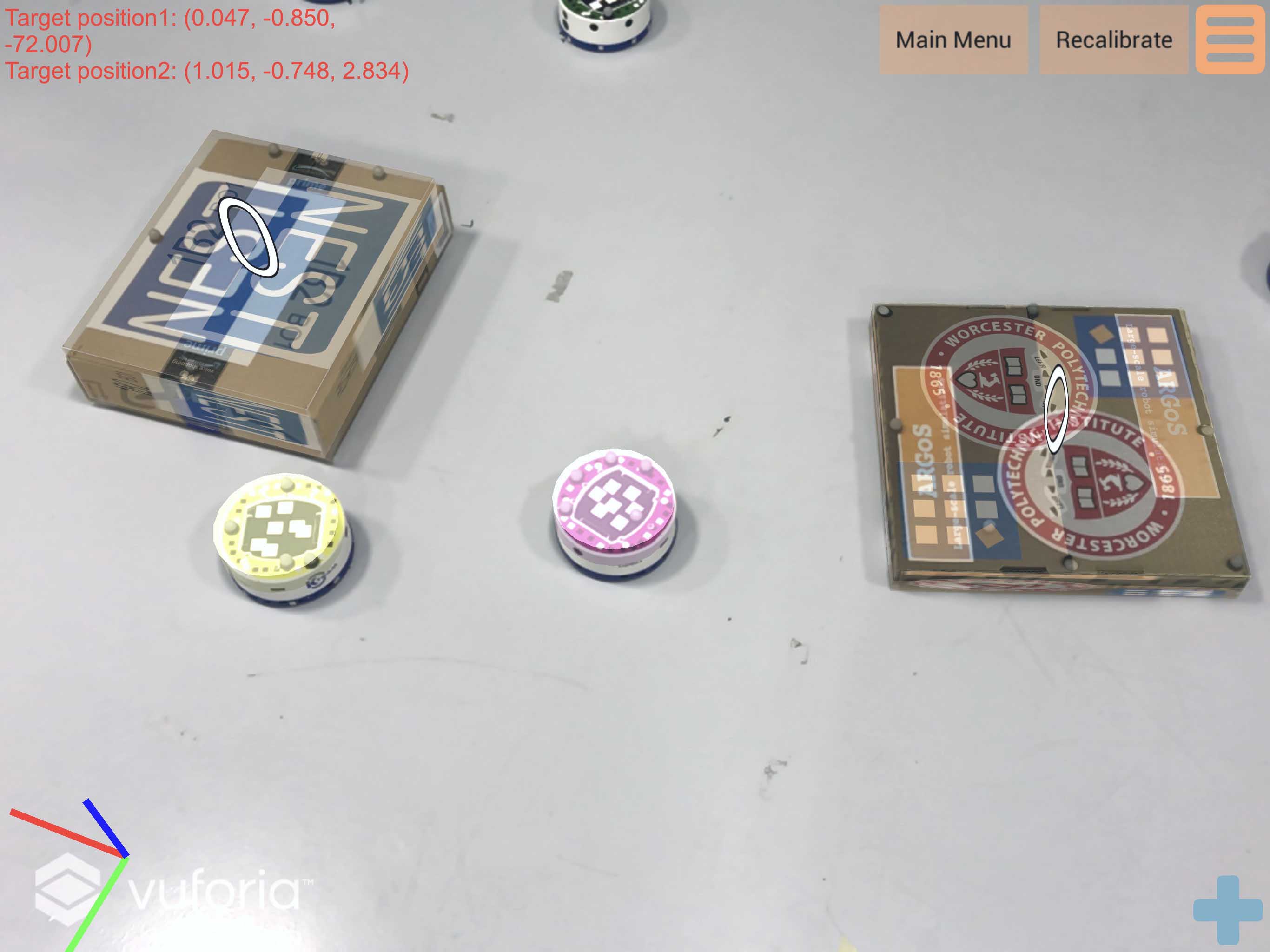}
    \caption{Two object tasks}
    \label{fig:modeT1}
  \end{subfigure}
  \begin{subfigure}[t]{0.2\textwidth}
    \includegraphics[width=\textwidth]{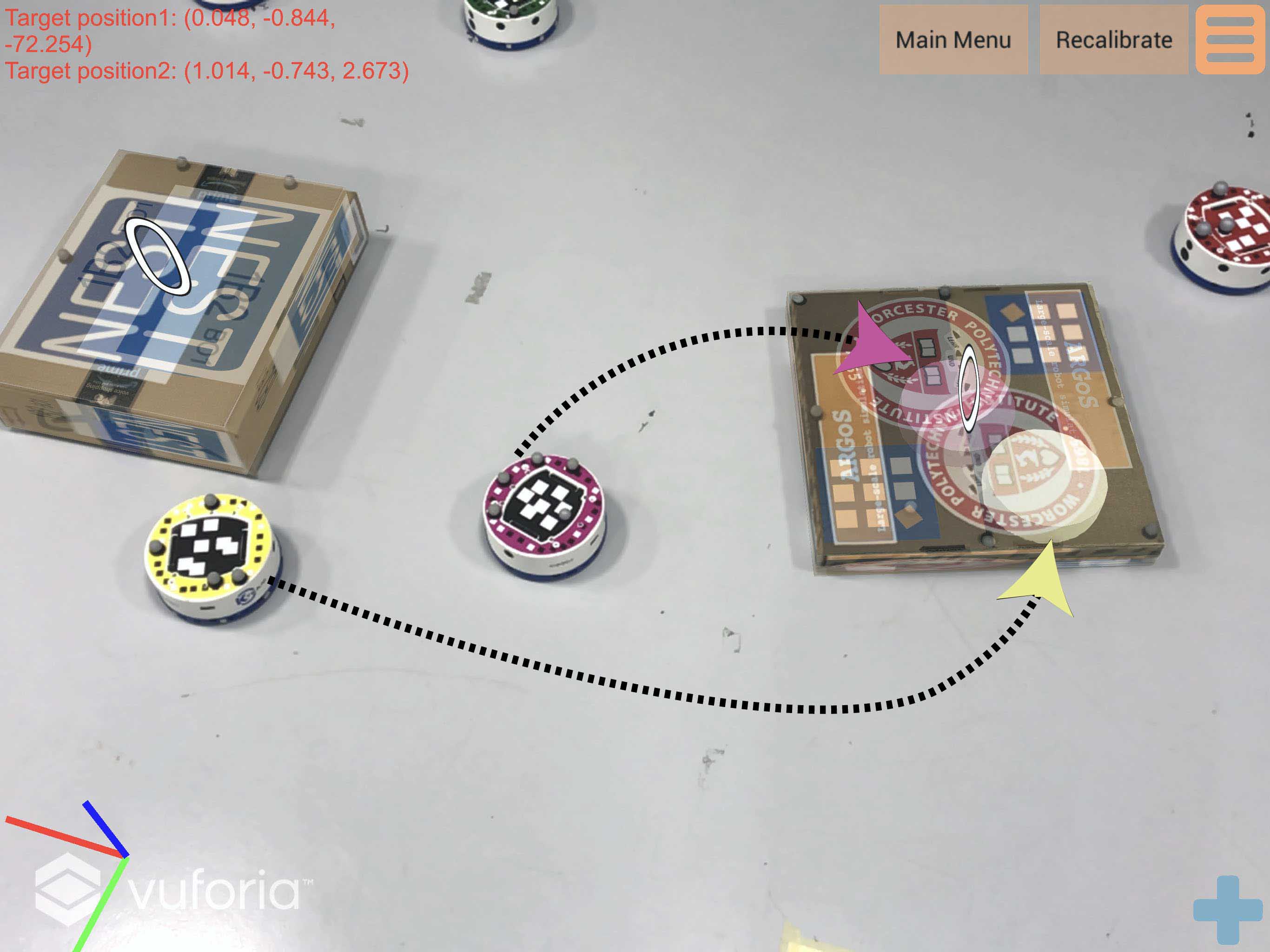}
    \caption{Team reassignment gesture}
    \label{fig:modeT2}
  \end{subfigure}
  \caption{Team reassignment through the interface to complete the task of
    moving an object. The overlaid dotted black arrow indicates the one-finger
    swipe gesture to move the virtual robot and the arrowhead color indicates
    the moved virtual robots.}\label{fig:modeT}
\end{figure}

\textbf{Team Reassignment.} A robot can be selected, and its virtual avatar can
be moved to overlap with the virtual object. This gesture assigns the robot to
the transport team corresponding to the object. This gesture is useful to
increase the resources dedicated to a heavy object, to replace broken/damaged
robots, and to distribute robot resources between multiple
operators. Fig.~\ref{fig:modeT1} shows the virtual robot overlapping the
physical robots. Fig.~\ref{fig:modeT2} demonstrates the reassignment of the
team.

\subsection{Shared Awareness}
The app broadcasts the modality changes performed by an operator. Other
hand-held devices receive these changes and reflect them in the augmented view,
thus making all the operators aware of the changes. The app shares this
information in real time, showing the virtual objects as they are manipulated by
other operators. This feature is useful to facilitate teamwork, to share
information on what an operator is currently controlling, and to avoid
conflicting control of a specific virtual object.

\subsection{Collective Transport}

\begin{figure}[t]
  \centering
  \includegraphics[width=0.4\textwidth]{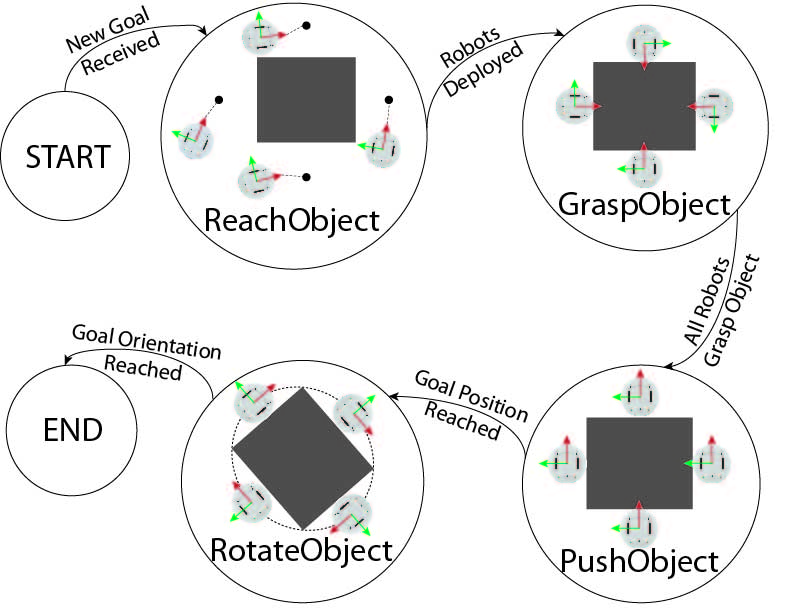}
  \caption{Collective transport state machine}
  \label{fig:statemachine}
\end{figure}
We employ a simple collective transport behavior based on the finite state
machine (FSM) shown in Fig.~\ref{fig:statemachine}. This behavior is identical
to the one presented in our previous work~\cite{patel2019}. We omit its full
description for reasons of brevity.






\section{User Study} \label{sec:userstudy}

\begin{figure}[t]
  \centering
  \begin{subfigure}[t]{0.2\textwidth}
    \includegraphics[width=\textwidth]{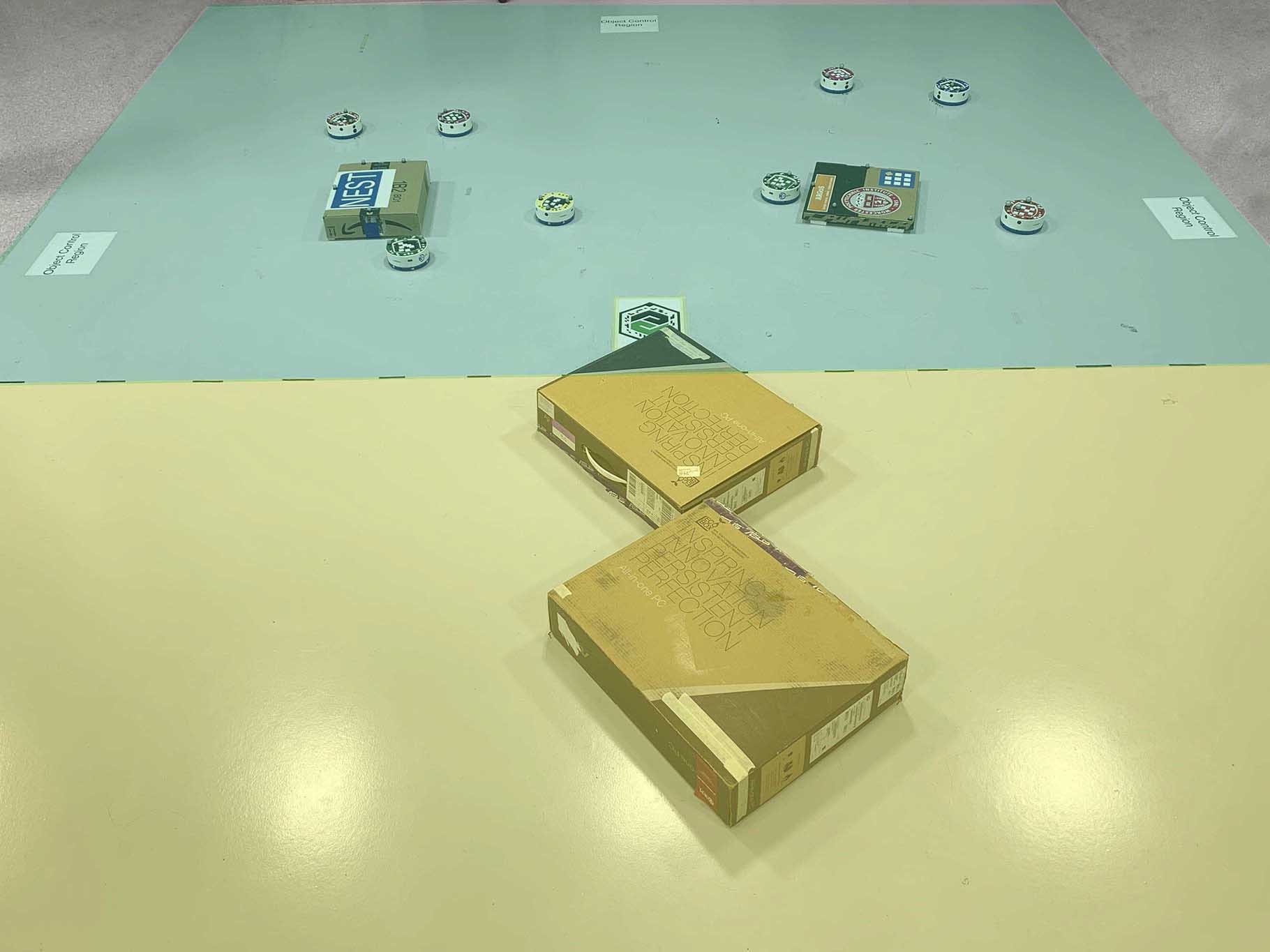}
    \caption{Initial positions}
    \label{fig:modeU1}
  \end{subfigure}
  \begin{subfigure}[t]{0.2\textwidth}
    \includegraphics[width=\textwidth]{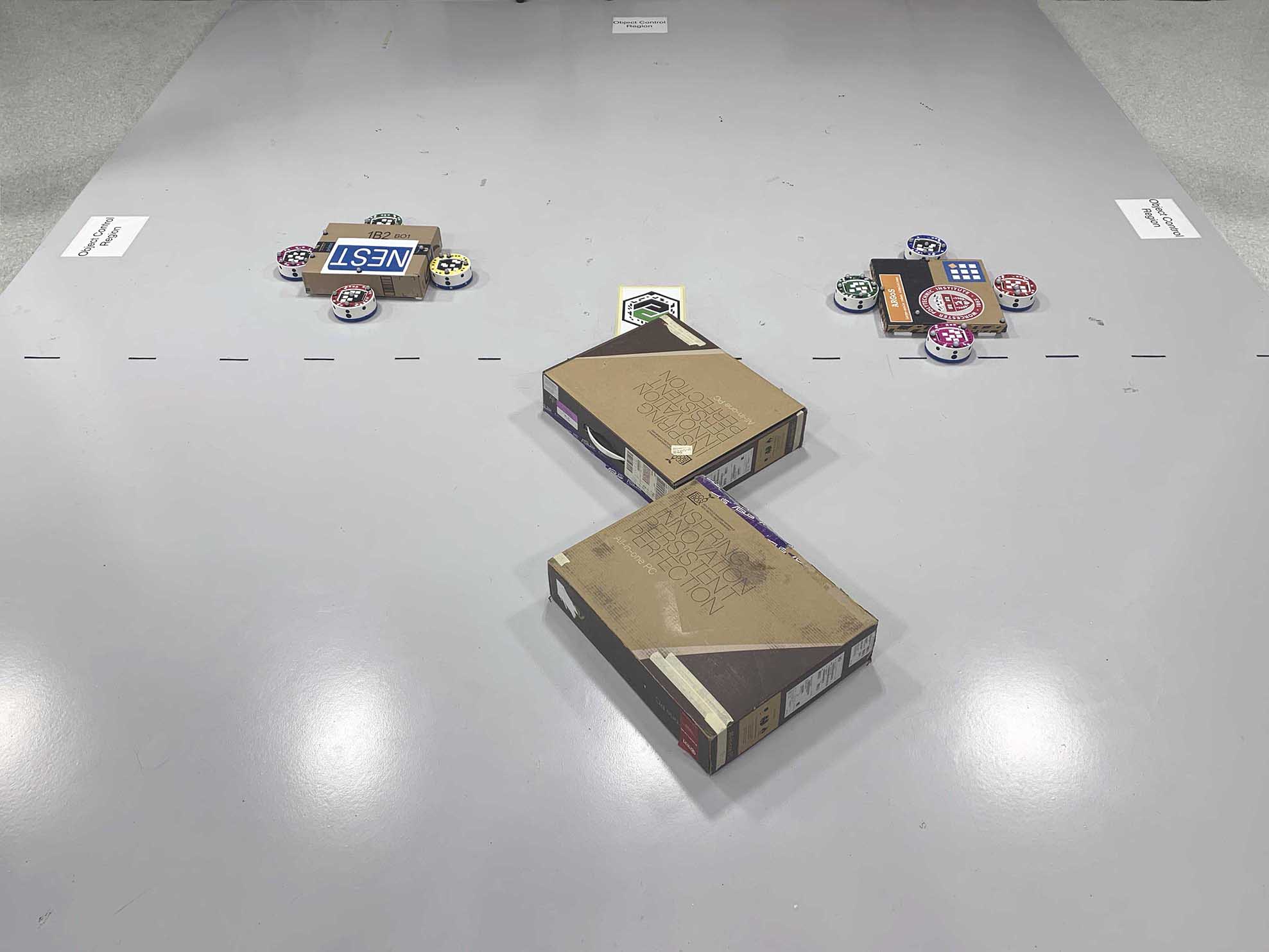}
    \caption{Object transport phase}
    \label{fig:modeU2}
  \end{subfigure}
  \begin{subfigure}[t]{0.2\textwidth}
    \includegraphics[width=\textwidth]{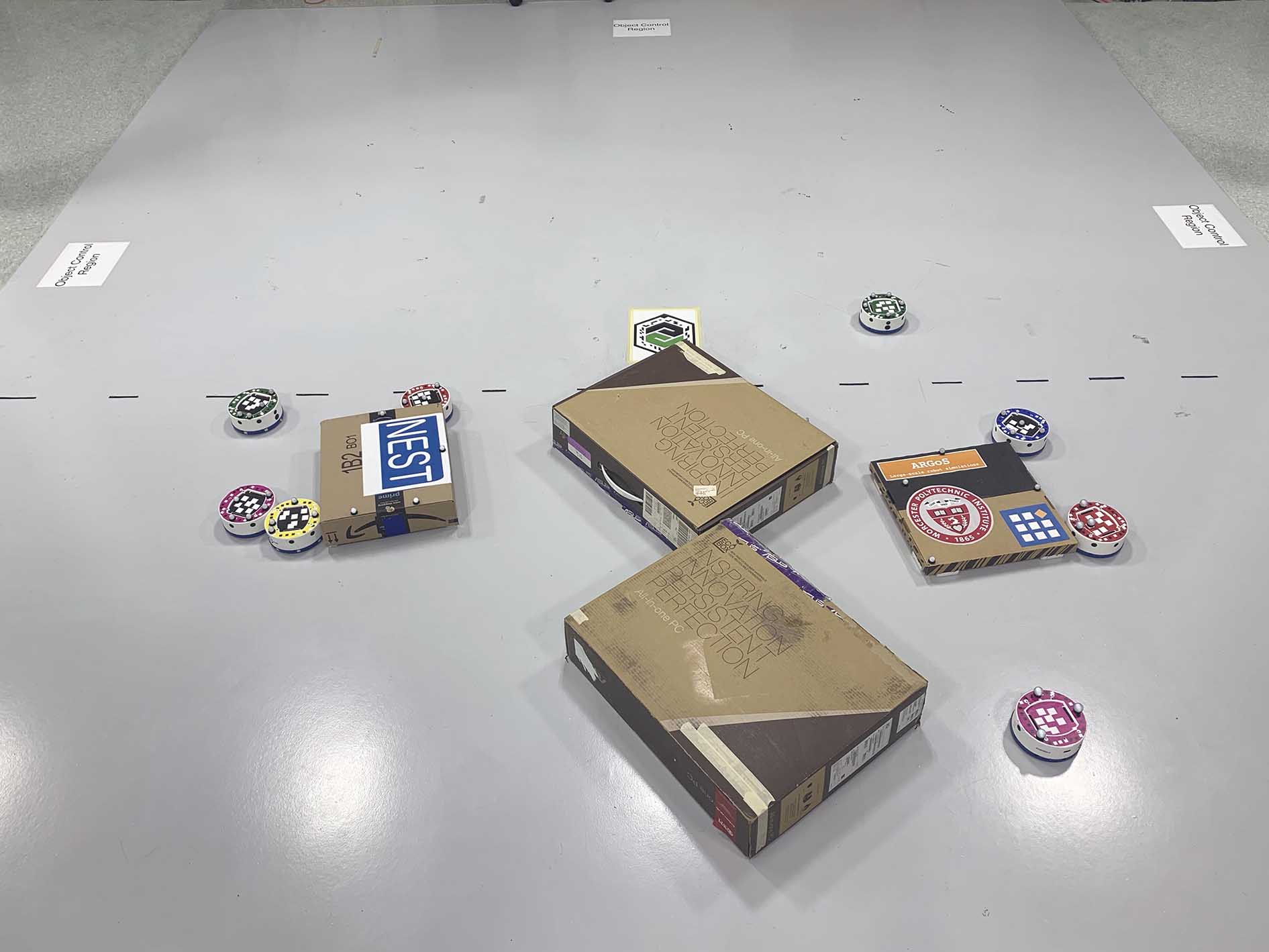}
    \caption{Robot control phase}
    \label{fig:modeU3}
  \end{subfigure}
  \begin{subfigure}[t]{0.2\textwidth}
    \includegraphics[width=\textwidth]{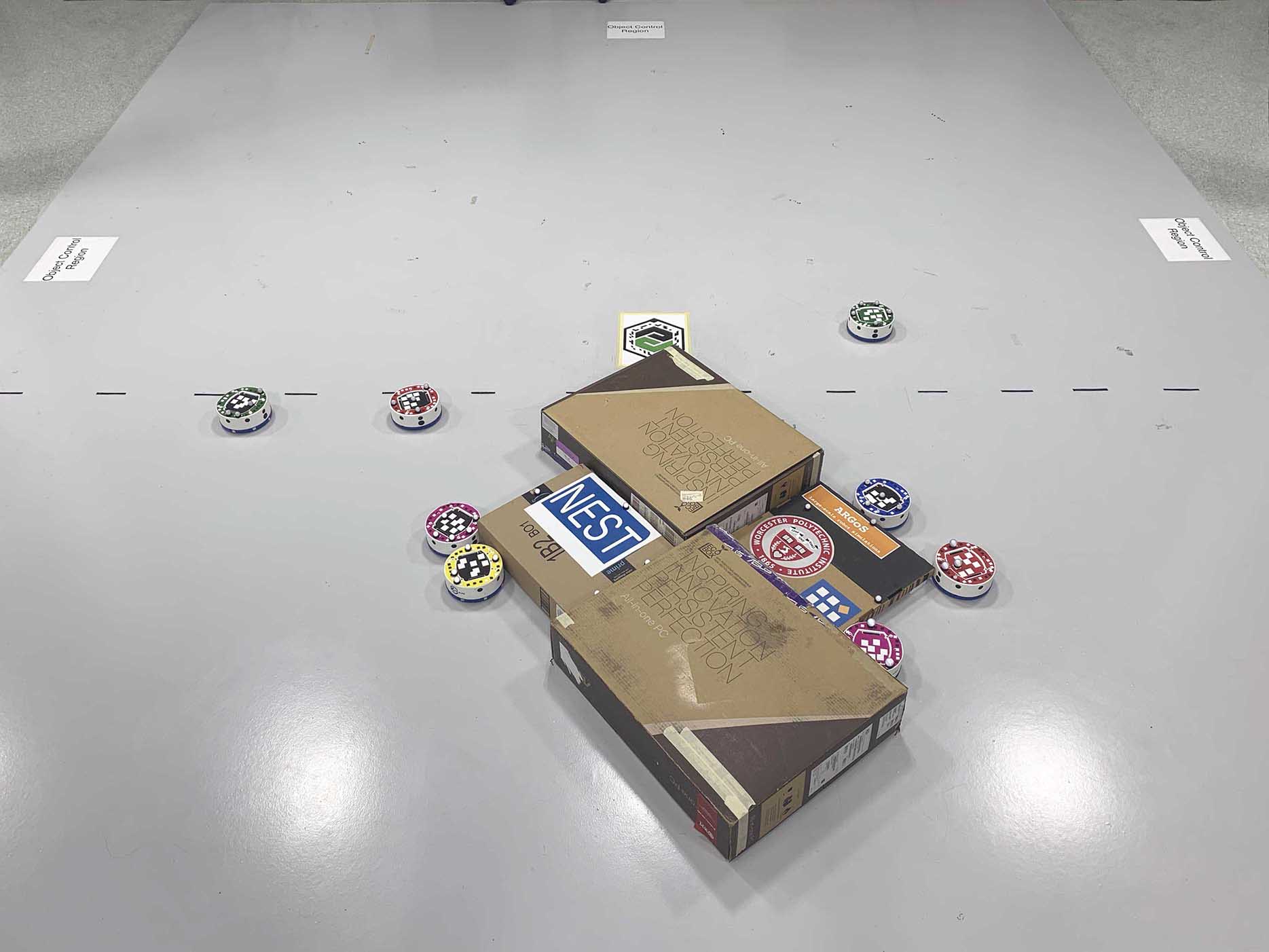}
    \caption{Desired structure}
    \label{fig:modeU4}
  \end{subfigure}
  \caption{Experimental setup of the user study. The overlaid green region
    indicates the transport region. The overlaid yellow region indicates the
    placement region.}\label{fig:modeU}
\end{figure} 

\subsection{Experimental Setup}
We designed a user study scenario in which two operators (O1 and O2) had to
supervise 8 robots in the construction of a simple structure. Because we focus
on the potential benefits of mixed granularity of control (MGOC), in our
experiments we considered two scenarios: one in which both operators used MGOC,
and one in which the operators were forced to use a single granularity of
control (SGOC).

\textbf{Phases.} Our construction scenario is composed of two phases. In Phase
1, the robots must transport an object in the general vicinity of its target
position. In Phase 2, the object must be pushed into its target position as
precisely as possible. For the task to be completed, the robots must place two
such objects. Fig.~\ref{fig:modeU1} shows the initial positions of the robots
and the objects in the field.

\textbf{Scenarios.} We considered two scenarios. In the \emph{MGOC
  scenario}, the operators are given the possibility to control the robots with
the full capabilities of the app. In addition, the operators are free to work in
any way they desire: they can work sequentially, collaborating on the first
object and then on the second; or they can work in parallel, focusing on
different objects. In contrast, in the \emph{SGOC scenario}, we established
specific roles and modalities of interaction for the operators. We divided the
field into two regions: the \emph{transport region} (corresponding to Phase 1)
and the \emph{placement region} (corresponding to Phase 2). We assigned a
specific operator to each region, and prevented operators from working outside
of their region: operator O1 was assigned to Phase 1 (transport), and operator
O2 was assigned to Phase 2 (placement). Motivated by the results of our previous
study~\cite{patel2019}, in the \emph{transport region} we allowed the operator
to only use object manipulation. On the other hand, in the \emph{placement
  region}, we allowed the operator to only use robot
control. Fig.~\ref{fig:modeU2} shows the collective transport behavior in the
\emph{transport region} and Fig.~\ref{fig:modeU3} shows the robots
controlled in the \emph{placement region}. Fig.~\ref{fig:modeU4} shows
the desired structure that the participants had to achieve for completing the
task. The dashed black line divides the field into the two regions.

\textbf{Procedure.} Each session of the study approximately took
75 minutes and involved two participants. The participants
first engaged in the two scenarios sequentially, and the order of the scenarios
was randomized to avoid any learning effects. At the beginning of each scenario,
we briefly explained to the participants the task they had to perform and gave
them 5 minutes to explore the app. The participants answered a
questionnaire after completing each scenario.

\textbf{Participants.}  We recruited a total of 28 students with ages ranging
from 21 to 30 years old ($\text{average} = 24.04 \pm 2.74$). None of them had
prior experience with the system.

\subsection{Metrics}
We recorded each participant's task activity metrics. In addition, we collected
the responses to the post-scenario questionnaire. We evaluated the following
metrics:

\textbf{Workload.} We employed the NASA TLX scale~\cite{hart_development_1988} 
for the participants to compare the workload during each scenarios. The participants
had to rank scenarios for each attribute of the scale. We used a Borda count~\cite{black1976partial} to 
find a winner based on the ranks assigned by the participants. We also recorded 
the number of user interactions (e.g., number of touches and gestures on the app).

\textbf{OOTL phenomenon.} We evaluated the OOTL performance problem by quantifying
situational awareness, which is the main factor for its
occurrence~\cite{endsley2017here}. To quantify situational awareness, we
employed the Situational Awareness Rating Technique
(SART)~\cite{selcon1991workload}. The participants had to rank the scenarios
for each attribute. We used a Borda count to determine the leading scenario 
based on the ranks assigned. Additionally, we recorded the activity
period (AP) of the participants during the scenario to analyze the total
duration of time they were active. We measured AP as the percentage of time a participant was interacting with the system.

\textbf{Trust.} We employed the group trust scale~\cite{allen2004exploring}
to analyze the trust between human teammates during a scenario, and the
human-robot trust sub-scale~\cite{schaefer2016measuring} to analyze the trust
in the robots' behavior.  The participants had to rank the scenarios based on these 
scales. We used a Borda count to determine the leading scenario.

\textbf{Task Performance.} To assess the overall performance of the system in
completing construction, we considered the time elapsed between the
start of a scenario and the moment in which the second object was placed in its
final destination.

\subsection{Results}

\begin{table}[t]
\centering
\caption{Borda count results of comparison study for workload based on NASA TLX scale attributes.
The gray cell indicate the leading scenario for each attribute. The mark $^-$ denotes negative scales. Lower ranking is better.}
\renewcommand{\arraystretch}{1.3}
\begin{tabular}{cc|c|c|c}
\hline\hline
                                        & \multicolumn{2}{|c}{O1}                         & \multicolumn{2}{|c}{O2} \\ \hline
\multicolumn{1}{c|}{NASA TLX Attributes}& SGOC       & MGOC                               & SGOC                               & MGOC       \\ \hline\hline
\multicolumn{1}{c|}{Mental Load$^-$}    & 15         & \cellcolor[HTML]{EFEFEF}27         & \cellcolor[HTML]{EFEFEF}23         & 19         \\ 
\multicolumn{1}{c|}{Physical Load$^-$}  & 18         & \cellcolor[HTML]{EFEFEF}24         & 21                                 & 21         \\ 
\multicolumn{1}{c|}{Temporal Load$^-$}  & 17         & \cellcolor[HTML]{EFEFEF}25         & 19                                 & \cellcolor[HTML]{EFEFEF}23         \\
\multicolumn{1}{c|}{Performance}        & 21         & 21                                 & 19                                 & \cellcolor[HTML]{EFEFEF}23         \\
\multicolumn{1}{c|}{Effort$^-$}         & 15         & \cellcolor[HTML]{EFEFEF}27         & 21                                 & 21         \\
\multicolumn{1}{c|}{Stress$^-$}         & 20         & \cellcolor[HTML]{EFEFEF}22         & 21                                 & 21         \\
\end{tabular}
\label{tab:workload}
\renewcommand{\arraystretch}{1}
\end{table}

\begin{table}[t]
\centering
\caption{Borda count results of comparison study for situational awareness based on SART scale attributes.
The gray cells indicate the leading scenario for each attribute.}
\renewcommand{\arraystretch}{1.3}
\begin{tabular}{cc|c|c|c}
\hline\hline
                                        & \multicolumn{2}{|c}{O1}                                            & \multicolumn{2}{|c}{O2} \\ \hline
\multicolumn{1}{c|}{SART Attributes}    & SGOC                          & MGOC                               & SGOC                               & MGOC       \\ \hline\hline
\multicolumn{1}{c|}{Complexity}         & 16                            & \cellcolor[HTML]{EFEFEF}26         & 20                                 & \cellcolor[HTML]{EFEFEF}22\\
\multicolumn{1}{c|}{Changeability}      & 18                            & \cellcolor[HTML]{EFEFEF}24         & 20                                 & \cellcolor[HTML]{EFEFEF}22\\ 
\multicolumn{1}{c|}{Variable}           & 16                            & \cellcolor[HTML]{EFEFEF}26         & 20                                 & \cellcolor[HTML]{EFEFEF}22\\ 
\multicolumn{1}{c|}{Arousal}            & 18                            & \cellcolor[HTML]{EFEFEF}24         & 21                                 & 21                        \\ 
\multicolumn{1}{c|}{Concentration}      & 16                            & \cellcolor[HTML]{EFEFEF}26         & \cellcolor[HTML]{EFEFEF}24         & 18                        \\ 
\multicolumn{1}{c|}{Mental Capacity}    & 15                            & \cellcolor[HTML]{EFEFEF}27         & \cellcolor[HTML]{EFEFEF}23         & 19                        \\ 
\multicolumn{1}{c|}{Information Gain}   & 19                            & \cellcolor[HTML]{EFEFEF}23         & \cellcolor[HTML]{EFEFEF}22         & 20                        \\ 
\multicolumn{1}{c|}{Familiarity}        & \cellcolor[HTML]{EFEFEF}23    & 19                                 & 19                                 & \cellcolor[HTML]{EFEFEF}23\\ 
\end{tabular}
\label{tab:awareness}
\renewcommand{\arraystretch}{1}
\end{table}

\begin{table}[t]
\centering
\caption{Borda count results of comparison study for trust based~\cite{allen2004exploring} and~\cite{schaefer2016measuring}.
The gray cells indicate the leading scenario for each attribute. The mark $^-$ denotes negative scales. Lower ranking is better.}
\renewcommand{\arraystretch}{1.3}
\begin{tabular}{cc|c|c|c}
\hline\hline
                                         & \multicolumn{2}{|c}{O1}                         & \multicolumn{2}{|c}{O2} \\ \hline
\multicolumn{1}{c|}{H-H Trust Attributes}& SGOC                          & MGOC                               & SGOC                               & MGOC       \\ \hline\hline
\multicolumn{1}{c|}{Honest}              & 17                            & \cellcolor[HTML]{EFEFEF}25         & 19                                 & \cellcolor[HTML]{EFEFEF}23         \\ 
\multicolumn{1}{c|}{Trustworthy}         & 18                            & \cellcolor[HTML]{EFEFEF}24         & 20                                 & \cellcolor[HTML]{EFEFEF}22         \\ 
\multicolumn{1}{c|}{Alert}               & 18                            & \cellcolor[HTML]{EFEFEF}24         & 20                                 & \cellcolor[HTML]{EFEFEF}22         \\
\multicolumn{1}{c|}{Help}                & 20                            & \cellcolor[HTML]{EFEFEF}22         & 16                                 & \cellcolor[HTML]{EFEFEF}26         \\
\multicolumn{1}{c|}{Will to Help}        & 17                            & \cellcolor[HTML]{EFEFEF}25         & 16                                 & \cellcolor[HTML]{EFEFEF}26         \\
\multicolumn{1}{c|}{Acceptance}          & 19                            & \cellcolor[HTML]{EFEFEF}23         & 17                                 & \cellcolor[HTML]{EFEFEF}25         \\\hline
\multicolumn{1}{c|}{H-R Trust Attributes}& SGOC                          & MGOC                               & SGOC                               & MGOC       \\ \hline\hline
\multicolumn{1}{c|}{Function Success}    & 21                            & 21                                 & 21                                 & 21         \\ 
\multicolumn{1}{c|}{Dependable}          & 18                            & \cellcolor[HTML]{EFEFEF}24         & 20                                 & \cellcolor[HTML]{EFEFEF}22         \\
\multicolumn{1}{c|}{Reliable}            & 19                            & \cellcolor[HTML]{EFEFEF}23         & 20                                 & \cellcolor[HTML]{EFEFEF}22         \\
\multicolumn{1}{c|}{Predictable}         & 21                            & 21                                 & 20                                 & \cellcolor[HTML]{EFEFEF}22         \\
\multicolumn{1}{c|}{Consistence}         & 20                            & \cellcolor[HTML]{EFEFEF}22         & 18                                 & \cellcolor[HTML]{EFEFEF}24         \\
\multicolumn{1}{c|}{Feedback}            & 19                            & \cellcolor[HTML]{EFEFEF}23         & 20                                 & \cellcolor[HTML]{EFEFEF}22         \\
\multicolumn{1}{c|}{Meet the Needs}      & 19                            & \cellcolor[HTML]{EFEFEF}23         & 20                                 & \cellcolor[HTML]{EFEFEF}22         \\
\multicolumn{1}{c|}{Provide Information} & \cellcolor[HTML]{EFEFEF}22    & 20                                 & 19                                 & \cellcolor[HTML]{EFEFEF}23         \\
\multicolumn{1}{c|}{Communication}       & 19                            & \cellcolor[HTML]{EFEFEF}23         & 20                                 & \cellcolor[HTML]{EFEFEF}22         \\
\multicolumn{1}{c|}{Performance}         & 19                            & \cellcolor[HTML]{EFEFEF}23         & 20                                 & \cellcolor[HTML]{EFEFEF}22         \\
\multicolumn{1}{c|}{Follow Directions}   & 20                            & \cellcolor[HTML]{EFEFEF}22         & 19                                 & \cellcolor[HTML]{EFEFEF}23         \\
\multicolumn{1}{c|}{Unresponsive$^-$}    & 20                            & \cellcolor[HTML]{EFEFEF}22         & \cellcolor[HTML]{EFEFEF}22         & 20         \\
\multicolumn{1}{c|}{Errors$^-$}          & 21                            & 21                                 & \cellcolor[HTML]{EFEFEF}23         & 19         \\
\multicolumn{1}{c|}{Malfunction$^-$}     & 21                            & 21                                 & \cellcolor[HTML]{EFEFEF}23         & 19         \\
\end{tabular}
\label{tab:trust}
\renewcommand{\arraystretch}{1}
\end{table}


\begin{figure}[t]
  \centering
  \includegraphics[width=0.29\textwidth]{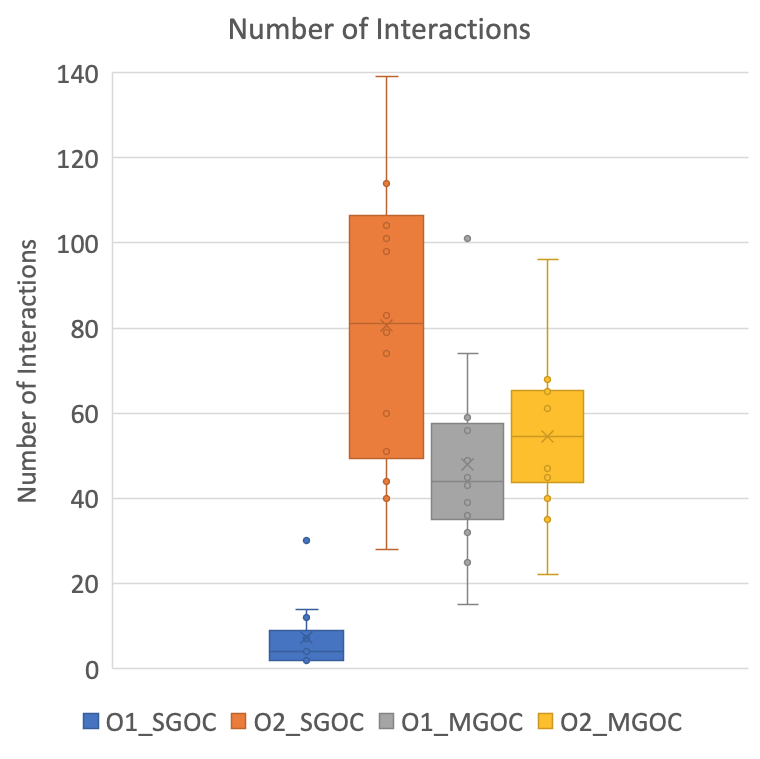}
  \caption{Number of interactions made with the hand-held device for both tasks
    by operators O1 and O2}
  \label{fig:interactions}
\end{figure}
\textbf{Workload.} Table~\ref{tab:workload}
reports the results of the workload comparison study for operators O1 and O2.
In the MGOC scenario, both O1 and O2 could choose how to interact
with the system and what to work on. In the SGOC scenario, O1 had to perform
transport with the object modality, while O2 was forced to perform placement
with the robot modality. The results show that, for O1, MGOC is much more
demanding than SGOC, while for O2 the workload in both scenarios is
approximately equal. These results are also confirmed in
Fig.~\ref{fig:interactions}, which reports the box plot for interactions
made with the hand-held device. Because the samples of the number of interactions in MGOC and SGOC are paired and non-parametric in nature, we use a Friedman test~\cite{friedman1937use} for statistical analysis. Setting a $p$-value of $0.01$ to establish statistical significance, we concluded that the difference in number of interactions is significant (Friedman test: $p = 0.0001$, $\chi^2 = 14.000$) 
between operators in SGOC, while the difference in number of interactions is not significant (Friedman test: $p = 0.109$, $\chi^2 = 2.571$) in MGOC.
These results indicate an imbalance in workload between operators in SGOC, leaving O1 out of the loop.


\begin{figure}[t]
  \centering
  \includegraphics[width=0.29\textwidth]{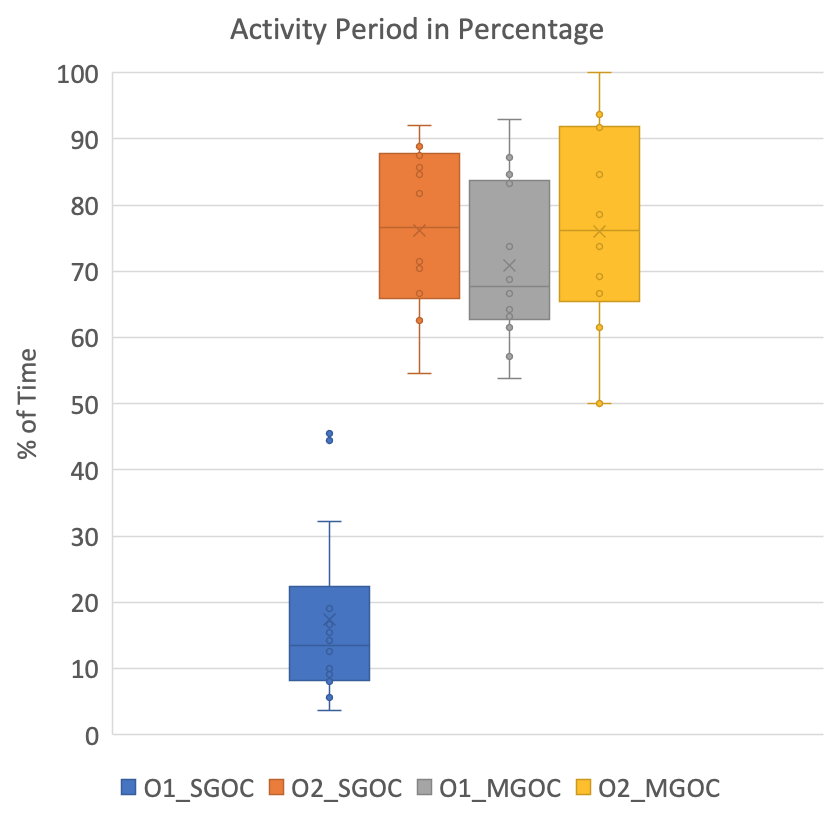}
  \caption{Activity Period in percentage of time for O1 and O2 during each scenario.}
  \label{fig:activity}
\end{figure}

\textbf{OOTL performance problem.} Table~\ref{tab:awareness} reports
the results of the situational awareness comparison study for O1 and O2. 
For O1, the SGOC scenario demands little attention; when
compared with MGOC, the data indicates that the latter results in a much higher
engagement of the operator in the task, while with SGOC the operators feel more out of the loop. In contrast, O2's levels of engagement and awareness are comparable
across the two scenarios. This interpretation is compatible with the data shown
in Fig.~\ref{fig:app}, which reports the percentage of the active period of both
operators in each scenario. Because the samples of the activity period data are paired and non-parametric in nature, we again used a Friedman test for statistical analysis. Setting a $p$-value of $0.01$ to establish statistical significance, we concluded that the difference in activity period 
is statistically significant between operators in SGOC (Friedman test: $p = 0.0001$, $\chi^2 = 14.000$), while in MGOC the difference in activity period is 
not (Friedman test: $p = 0.248$, $\chi^2 = 1.333$).


\textbf{Trust.} Table~\ref{tab:trust} reports the results of our trust comparison study
for O1 and O2. Both O1 and O2 reported higher trust in MGOC than in SGOC. 

\begin{figure}[t]
  \centering
  \includegraphics[width=0.29\textwidth]{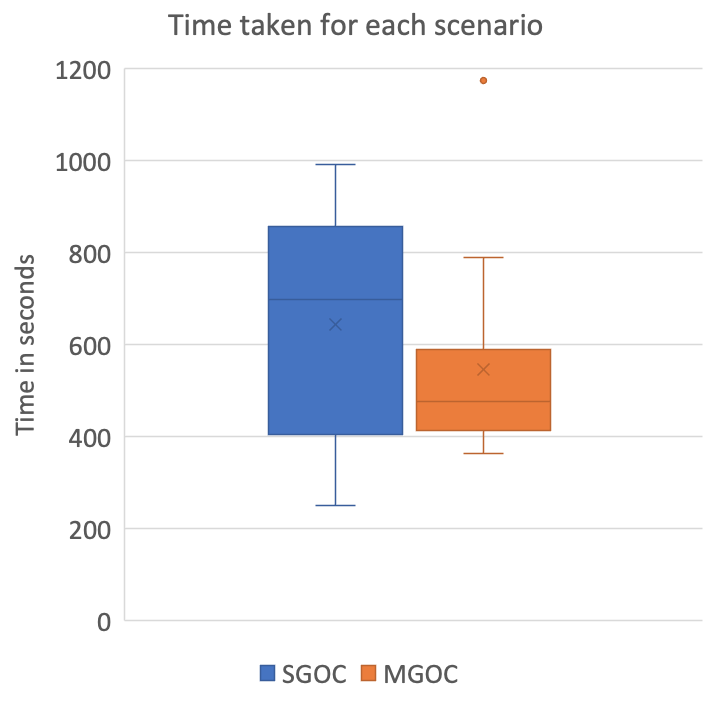}
  \caption{Performance recorded in terms of time taken to complete each task.}
  \label{fig:performance}
\end{figure}
\textbf{Task Performance.} Fig.~\ref{fig:performance} shows a box plot of task
performance for SGOC and MGOC. Because the samples of the performance data are paired and non-parametric 
in nature, we again used a Friedman test for statistical analysis.
Setting a $p$-value of $0.05$ to establish statistical significance, we concluded that the 
difference in activity period is not statistically significant between SGOC and MGOC 
(Friedman test: $p = 0.109$, $\chi^2 = 2.571$).
The median completion times we observed were
\unit[10.63]{min} and \unit[7.93]{min} for SGOC and MGOC, respectively. 
The median suggests that MGOC outperforms SGOC in terms of completion time.

\section{Discussion}
\label{sec:discussion}

The results of our user study allow us to draw a number of interesting
conclusions about the nature of multi-human multi-robot interaction.

First, the case in which the operators are given an equal role (MGOC scenario)
corresponded with the best system performance. While we never suggested to the
operators how to structure their work, all the operators pairs quickly settled
on working in parallel on both phases. Often, one operator completed her task
faster than the other---in this case, the faster operator switched to help the
slower one. This resulted in both operators being constantly engaged in the task
and also in a better sense of mutual trust and teamwork.

In contrast, when we forcefully assigned specific roles and modalities of
interaction, we found that the operator with a lower workload became easily
distracted from the task. It is important to notice that, in Phase 1, the most
efficient option to complete the task is object manipulation, while in Phase 2
it is robot manipulation. These modalities were also preferred in the MGOC
scenarios when the operators acted in parallel. Therefore the only significant
difference between SGOC and MGOC was in the forced role assignment.

Our results are consistent with~\cite{endsley2017here} and suggest that, when
distributing control responsibility across operators, the OOTL problem
affects the system performance. The operators prefer to reach their limit of
apprehension and remain engaged in the system, rather than having long periods
of inactivity followed by sudden moment of high load. In addition, the latter 
scenario might prove stressful because of the difficulty of catching up quickly 
when a system composed of many parts presents itself in an unknown state that 
demands attention.

As a consequence, specialization across operators (the SGOC scenario) is not
necessarily the best option. In designing the roles of the operators of a
multi-robot system, special attention must be paid to balancing the workload,
keeping engagement high, and allowing for a healthy level of overlap across
operators, to foster the kind of teamwork we observed in the MGOC case. In our
user study, 25 out of 28 participants reported that they prefer the MGOC
scenario over the SGOC one. Different tasks, robot behaviors, and applications
might reveal more complex phenomena. More research is required to understand the
connection between these issues and the challenge of increasing human
performance in an automated system.

\section{Conclusion and Future Work} \label{sec:conclusion}

In this paper, we study the role of the out-of-the-loop (OOTL) phenomenon in the
context of multi-human multi-robot interaction. Our paper offers two main
contributions.

Our technological contribution is the first collaborative augmented-reality app
that allows for mixed-granularity control---including environment-oriented,
team-oriented, and robot-oriented control modalities. Using this app, we
conducted a user study involving 28 participants and 8 real robots to study how
aspects such as role assignment and modalities of interaction affect the
engagement of the operators and, ultimately, the performance of the overall
system.

Our scientific contribution consists in the insight that, when establishing the
responsibilities of multiple operators, specialization might not be the most
desirable option. This is because a certain degree of responsibility overlap
across the operators might offer flexibility, resulting in an increased sense of
mutual trust among operators. In addition, a more balanced workload across
operators prevents the insurgence of OOTL phenomena and improve the system
performance.

Future work will aim to understand the effects of \textit{interface training} on
task performance~\cite{you2016curiosity}, for instance, giving more time to
understand and test the interface before participating in the task. Additionally, we
will aim at making the relationship between humans and robots more legible and
\textit{transparent}~\cite{roundtree2019transparency} and understand ways to
keep the operators engaged in the system.

\section*{Acknowledgements}
This work was funded by a grant from mRobot Technology Co, Shanghai, China.

\bibliographystyle{IEEEtran}
\bibliography{ref}

\begin{thebibliography}{10}
\providecommand{\url}[1]{#1}
\csname url@samestyle\endcsname
\providecommand{\newblock}{\relax}
\providecommand{\bibinfo}[2]{#2}
\providecommand{\BIBentrySTDinterwordspacing}{\spaceskip=0pt\relax}
\providecommand{\BIBentryALTinterwordstretchfactor}{4}
\providecommand{\BIBentryALTinterwordspacing}{\spaceskip=\fontdimen2\font plus
\BIBentryALTinterwordstretchfactor\fontdimen3\font minus
  \fontdimen4\font\relax}
\providecommand{\BIBforeignlanguage}[2]{{%
\expandafter\ifx\csname l@#1\endcsname\relax
\typeout{** WARNING: IEEEtran.bst: No hyphenation pattern has been}%
\typeout{** loaded for the language `#1'. Using the pattern for}%
\typeout{** the default language instead.}%
\else
\language=\csname l@#1\endcsname
\fi
#2}}
\providecommand{\BIBdecl}{\relax}
\BIBdecl

\bibitem{Brambilla2013}
M.~Brambilla, E.~Ferrante, M.~Birattari, and M.~Dorigo, ``{Swarm robotics: A
  review from the swarm engineering perspective},'' \emph{Swarm Intelligence},
  vol.~7, no.~1, pp. 1--41, 2013.

\bibitem{gates2007robot}
B.~Gates, ``A robot in every home,'' \emph{Scientific American}, vol. 296,
  no.~1, pp. 58--65, 2007.

\bibitem{rubio2012mining}
R.~F. Rubio, ``Mining: the challenge knocks on our door,'' \emph{Mine Water and
  the Environment}, vol.~31, no.~1, pp. 69--73, 2012.

\bibitem{oh2009bridge}
J.-K. Oh, G.~Jang, S.~Oh, J.~H. Lee, B.-J. Yi, Y.~S. Moon, J.~S. Lee, and
  Y.~Choi, ``Bridge inspection robot system with machine vision,''
  \emph{Automation in Construction}, vol.~18, no.~7, pp. 929--941, 2009.

\bibitem{murphy2014disaster}
R.~R. Murphy, \emph{Disaster robotics}.\hskip 1em plus 0.5em minus 0.4em\relax
  MIT press, 2014.

\bibitem{hamins2015research}
A.~P. Hamins, N.~P. Bryner, A.~W. Jones, G.~H. Koepke \emph{et~al.}, ``Research
  roadmap for smart fire fighting,'' Tech. Rep., 2015.

\bibitem{goldsmith1999book}
D.~Goldsmith, ``Book review: Voyage to the milky way: the future of space
  exploration/tv books, 1999,'' \emph{Sky and Telescope}, vol.~98, no.~5,
  p.~81, 1999.

\bibitem{kolling2015human}
A.~Kolling, P.~Walker, N.~Chakraborty, K.~Sycara, and M.~Lewis, ``Human
  interaction with robot swarms: A survey,'' \emph{IEEE Transactions on
  Human-Machine Systems}, vol.~46, no.~1, pp. 9--26, 2015.

\bibitem{miller1956magical}
G.~A. Miller, ``The magical number seven, plus or minus two: Some limits on our
  capacity for processing information.'' \emph{Psychological review}, vol.~63,
  no.~2, p.~81, 1956.

\bibitem{lewis2010choosing}
M.~Lewis, H.~Wang, S.~Y. Chien, P.~Velagapudi, P.~Scerri, and K.~Sycara,
  ``Choosing autonomy modes for multirobot search,'' \emph{Human Factors},
  vol.~52, no.~2, pp. 225--233, 2010.

\bibitem{endsley2017here}
M.~R. Endsley, ``From here to autonomy: lessons learned from human--automation
  research,'' \emph{Human factors}, vol.~59, no.~1, pp. 5--27, 2017.

\bibitem{hussein2018mixed}
A.~Hussein and H.~Abbass, ``Mixed initiative systems for human-swarm
  interaction: Opportunities and challenges,'' in \emph{2018 2nd Annual Systems
  Modelling Conference (SMC)}.\hskip 1em plus 0.5em minus 0.4em\relax IEEE,
  2018, pp. 1--8.

\bibitem{chen2014human}
J.~Y. Chen and M.~J. Barnes, ``Human--agent teaming for multirobot control: A
  review of human factors issues,'' \emph{IEEE Transactions on Human-Machine
  Systems}, vol.~44, no.~1, pp. 13--29, 2014.

\bibitem{allen2004exploring}
K.~Allen, R.~Bergin, and K.~Pickar, ``Exploring trust, group satisfaction, and
  performance in geographically dispersed and co-located university technology
  commercialization teams,'' in \emph{VentureWell. Proceedings of Open, the
  Annual Conference}.\hskip 1em plus 0.5em minus 0.4em\relax National
  Collegiate Inventors \& Innovators Alliance, 2004, p. 201.

\bibitem{mcbride2011understanding}
S.~E. McBride, W.~A. Rogers, and A.~D. Fisk, ``Understanding the effect of
  workload on automation use for younger and older adults,'' \emph{Human
  factors}, vol.~53, no.~6, pp. 672--686, 2011.

\bibitem{riley2005situation}
J.~M. Riley and M.~R. Endsley, ``Situation awareness in hri with collaborating
  remotely piloted vehicles,'' in \emph{proceedings of the Human Factors and
  Ergonomics Society Annual Meeting}, vol.~49, no.~3.\hskip 1em plus 0.5em
  minus 0.4em\relax SAGE Publications Sage CA: Los Angeles, CA, 2005, pp.
  407--411.

\bibitem{lee2008review}
J.~D. Lee, ``Review of a pivotal human factors article:“humans and
  automation: use, misuse, disuse, abuse”,'' \emph{Human Factors}, vol.~50,
  no.~3, pp. 404--410, 2008.

\bibitem{parasuraman1997humans}
R.~Parasuraman and V.~Riley, ``Humans and automation: Use, misuse, disuse,
  abuse,'' \emph{Human factors}, vol.~39, no.~2, pp. 230--253, 1997.

\bibitem{endsley1995out}
M.~R. Endsley and E.~O. Kiris, ``The out-of-the-loop performance problem and
  level of control in automation,'' \emph{Human factors}, vol.~37, no.~2, pp.
  381--394, 1995.

\bibitem{gouraud2017autopilot}
J.~Gouraud, A.~Delorme, and B.~Berberian, ``Autopilot, mind wandering, and the
  out of the loop performance problem,'' \emph{Frontiers in neuroscience},
  vol.~11, p. 541, 2017.

\bibitem{patel2019}
J.~Patel, Y.~Xu, and C.~Pinciroli, ``Mixed-granularity human-swarm
  interaction,'' in \emph{Robotics and {Automation} ({ICRA}), 2019 {IEEE}
  {International} {Conference} on}.\hskip 1em plus 0.5em minus 0.4em\relax
  IEEE, 2019.

\bibitem{lewis_effects_2011}
M.~Lewis and K.~Sycara, ``Effects of automation on situation awareness in
  controlling robot teams,'' p.~7.

\bibitem{you2016curiosity}
S.~You and L.~Robert~Jr, ``Curiosity vs. control: Impacts of training on
  performance of teams working with robots,'' in \emph{Proceedings of the 19th
  ACM Conference on Computer Supported Cooperative Work and Social Computing
  Companion}.\hskip 1em plus 0.5em minus 0.4em\relax ACM, 2016, pp. 449--452.

\bibitem{lewis_teams_2010}
\BIBentryALTinterwordspacing
M.~Lewis, H.~Wang, {Shih-Yi Chien}, P.~Scerri, P.~Velagapudi, K.~Sycara, and
  B.~Kane, ``Teams organization and performance in multi-human/multi-robot
  teams,'' in \emph{2010 {IEEE} International Conference on Systems, Man and
  Cybernetics}.\hskip 1em plus 0.5em minus 0.4em\relax {IEEE}, pp. 1617--1623.
  [Online]. Available: \url{http://ieeexplore.ieee.org/document/5642379/}
\BIBentrySTDinterwordspacing

\bibitem{gromov_wearable_2016}
\BIBentryALTinterwordspacing
B.~Gromov, L.~M. Gambardella, and G.~A. Di~Caro,
  ``\BIBforeignlanguage{en}{Wearable multi-modal interface for human
  multi-robot interaction}.''\hskip 1em plus 0.5em minus 0.4em\relax IEEE, Oct.
  2016, pp. 240--245. [Online]. Available:
  \url{http://ieeexplore.ieee.org/document/7784305/}
\BIBentrySTDinterwordspacing

\bibitem{kapellmann-zafra_human-robot_2016}
\BIBentryALTinterwordspacing
G.~Kapellmann-Zafra, N.~Salomons, A.~Kolling, and R.~Gro{\ss}, ``Human-{Robot}
  {Swarm} {Interaction} with {Limited} {Situational} {Awareness},'' in
  \emph{International {Conference} on {Swarm} {Intelligence}}.\hskip 1em plus
  0.5em minus 0.4em\relax Springer, 2016, pp. 125--136. [Online]. Available:
  \url{http://link.springer.com/chapter/10.1007/978-3-319-44427-7_11}
\BIBentrySTDinterwordspacing

\bibitem{alonso-mora_gesture_2015}
\BIBentryALTinterwordspacing
J.~Alonso-Mora, S.~Haegeli~Lohaus, P.~Leemann, R.~Siegwart, and P.~Beardsley,
  ``\BIBforeignlanguage{en}{Gesture based human - {Multi}-robot swarm
  interaction and its application to an interactive display},'' in
  \emph{\BIBforeignlanguage{en}{2015 {IEEE} {International} {Conference} on
  {Robotics} and {Automation} ({ICRA})}}.\hskip 1em plus 0.5em minus
  0.4em\relax Seattle, WA, USA: IEEE, May 2015, pp. 5948--5953. [Online].
  Available: \url{http://ieeexplore.ieee.org/document/7140033/}
\BIBentrySTDinterwordspacing

\bibitem{nagi_human-swarm_2014}
\BIBentryALTinterwordspacing
J.~Nagi, A.~Giusti, L.~M. Gambardella, and G.~A. Di~Caro,
  ``\BIBforeignlanguage{en}{Human-swarm interaction using spatial gestures},''
  in \emph{\BIBforeignlanguage{en}{2014 {IEEE}/{RSJ} {International}
  {Conference} on {Intelligent} {Robots} and {Systems}}}.\hskip 1em plus 0.5em
  minus 0.4em\relax Chicago, IL, USA: IEEE, Sep. 2014, pp. 3834--3841.
  [Online]. Available: \url{http://ieeexplore.ieee.org/document/6943101/}
\BIBentrySTDinterwordspacing

\bibitem{natraj_gesturing_2014}
\BIBentryALTinterwordspacing
G.~Podevijn, R.~O’Grady, Y.~S.~G. Nashed, and M.~Dorigo,
  ``\BIBforeignlanguage{en}{Gesturing at {Subswarms}: {Towards} {Direct}
  {Human} {Control} of {Robot} {Swarms}},'' in
  \emph{\BIBforeignlanguage{en}{Towards {Autonomous} {Robotic} {Systems}}},
  A.~Natraj, S.~Cameron, C.~Melhuish, and M.~Witkowski, Eds.\hskip 1em plus
  0.5em minus 0.4em\relax Berlin, Heidelberg: Springer Berlin Heidelberg, 2014,
  vol. 8069, pp. 390--403. [Online]. Available:
  \url{http://link.springer.com/10.1007/978-3-662-43645-5_41}
\BIBentrySTDinterwordspacing

\bibitem{nagavalli2017multi}
S.~Nagavalli, M.~Chandarana, K.~Sycara, and M.~Lewis, ``Multi-operator gesture
  control of robotic swarms using wearable devices,'' in \emph{Proceedings of
  the Tenth International Conference on Advances in Computer-Human
  Interactions}.\hskip 1em plus 0.5em minus 0.4em\relax IARIA, 2017.

\bibitem{bashyal_human_2008}
\BIBentryALTinterwordspacing
S.~Bashyal and G.~K. Venayagamoorthy, ``Human swarm interaction for radiation
  source search and localization.''\hskip 1em plus 0.5em minus 0.4em\relax
  IEEE, Sep. 2008, pp. 1--8. [Online]. Available:
  \url{http://ieeexplore.ieee.org/document/4668287/}
\BIBentrySTDinterwordspacing

\bibitem{diaz-mercado_distributed_2015}
Y.~Diaz-Mercado, S.~G. Lee, and M.~Egerstedt, ``Distributed dynamic density
  coverage for human-swarm interactions,'' in \emph{American {Control}
  {Conference} ({ACC}), 2015}.\hskip 1em plus 0.5em minus 0.4em\relax IEEE,
  2015, pp. 353--358.

\bibitem{kolling_human_2013}
\BIBentryALTinterwordspacing
A.~Kolling, K.~Sycara, S.~Nunnally, and M.~Lewis,
  ``\BIBforeignlanguage{en}{Human {Swarm} {Interaction}: {An} {Experimental}
  {Study} of {Two} {Types} of {Interaction} with {Foraging} {Swarms}},''
  \emph{\BIBforeignlanguage{en}{Journal of Human-Robot Interaction}}, vol.~2,
  no.~2, Jun. 2013. [Online]. Available:
  \url{http://dl.acm.org/citation.cfm?id=3109714}
\BIBentrySTDinterwordspacing

\bibitem{ayanian_controlling_2014}
\BIBentryALTinterwordspacing
N.~Ayanian, A.~Spielberg, M.~Arbesfeld, J.~Strauss, and D.~Rus, ``Controlling a
  team of robots with a single input,'' in \emph{Robotics and {Automation}
  ({ICRA}), 2014 {IEEE} {International} {Conference} on}.\hskip 1em plus 0.5em
  minus 0.4em\relax IEEE, 2014, pp. 1755--1762. [Online]. Available:
  \url{http://ieeexplore.ieee.org/abstract/document/6907088/}
\BIBentrySTDinterwordspacing

\bibitem{johnson2008human}
M.~Johnson, P.~J. Feltovich, J.~M. Bradshaw, and L.~Bunch, ``Human-robot
  coordination through dynamic regulation,'' in \emph{2008 IEEE International
  Conference on Robotics and Automation}.\hskip 1em plus 0.5em minus
  0.4em\relax IEEE, 2008, pp. 2159--2164.

\bibitem{dias_dynamically_2006}
M.~B. Dias, ``Dynamically formed human-robot teams performing coordinated
  tasks,'' p.~9.

\bibitem{malvankar-mehta_optimal_2015}
\BIBentryALTinterwordspacing
M.~S. Malvankar-Mehta and S.~S. Mehta, ``Optimal task allocation in multi-human
  multi-robot interaction,'' vol.~9, no.~8, pp. 1787--1803. [Online].
  Available: \url{http://link.springer.com/10.1007/s11590-015-0890-7}
\BIBentrySTDinterwordspacing

\bibitem{huang_human-oriented_2010}
K.-C. Huang, J.-Y. Li, and L.-C. Fu, ``\BIBforeignlanguage{en}{Human-oriented
  navigation for service providing in home environment},''
  \emph{\BIBforeignlanguage{en}{Proceedings of SICE Annual Conference 2010}},
  p.~6, 2010.

\bibitem{claes_multi_2018}
\BIBentryALTinterwordspacing
D.~Claes and K.~Tuyls, ``\BIBforeignlanguage{en}{Multi robot collision
  avoidance in a shared workspace},'' \emph{\BIBforeignlanguage{en}{Autonomous
  Robots}}, vol.~42, no.~8, pp. 1749--1770, Dec. 2018. [Online]. Available:
  \url{http://link.springer.com/10.1007/s10514-018-9726-5}
\BIBentrySTDinterwordspacing

\bibitem{wang_trust-based_2018}
\BIBentryALTinterwordspacing
Y.~Wang, L.~R. Humphrey, Z.~Liao, and H.~Zheng,
  ``\BIBforeignlanguage{en}{Trust-based {Multi}-{Robot} {Symbolic} {Motion}
  {Planning} with a {Human}-in-the-{Loop}},'' \emph{\BIBforeignlanguage{en}{ACM
  Transactions on Interactive Intelligent Systems}}, vol.~8, no.~4, pp. 1--33,
  Nov. 2018, arXiv: 1808.05120. [Online]. Available:
  \url{http://arxiv.org/abs/1808.05120}
\BIBentrySTDinterwordspacing

\bibitem{tseng_multi-human_2014}
\BIBentryALTinterwordspacing
S.-H. Tseng, Y.-H. Hsu, Y.-S. Chiang, T.-Y. Wu, and L.-C. Fu,
  ``\BIBforeignlanguage{en}{Multi-human spatial social pattern understanding
  for a multi-modal robot through nonverbal social signals},'' in
  \emph{\BIBforeignlanguage{en}{The 23rd {IEEE} {International} {Symposium} on
  {Robot} and {Human} {Interactive} {Communication}}}.\hskip 1em plus 0.5em
  minus 0.4em\relax Edinburgh, UK: IEEE, Aug. 2014, pp. 531--536. [Online].
  Available: \url{http://ieeexplore.ieee.org/document/6926307/}
\BIBentrySTDinterwordspacing

\bibitem{higuera_socially-driven_2012}
\BIBentryALTinterwordspacing
J.~C.~G. Higuera, A.~Xu, F.~Shkurti, and G.~Dudek,
  ``\BIBforeignlanguage{en}{Socially-{Driven} {Collective} {Path} {Planning}
  for {Robot} {Missions}},'' in \emph{\BIBforeignlanguage{en}{2012 {Ninth}
  {Conference} on {Computer} and {Robot} {Vision}}}.\hskip 1em plus 0.5em minus
  0.4em\relax Toronto, Ontario, Canada: IEEE, May 2012, pp. 417--424. [Online].
  Available: \url{http://ieeexplore.ieee.org/document/6233171/}
\BIBentrySTDinterwordspacing

\bibitem{bajcsy_scalable_2018}
\BIBentryALTinterwordspacing
A.~Bajcsy, S.~L. Herbert, D.~Fridovich-Keil, J.~F. Fisac, S.~Deglurkar, A.~D.
  Dragan, and C.~J. Tomlin, ``\BIBforeignlanguage{en}{A {Scalable} {Framework}
  {For} {Real}-{Time} {Multi}-{Robot}, {Multi}-{Human} {Collision}
  {Avoidance}},'' \emph{\BIBforeignlanguage{en}{arXiv:1811.05929 [cs]}}, Nov.
  2018, arXiv: 1811.05929. [Online]. Available:
  \url{http://arxiv.org/abs/1811.05929}
\BIBentrySTDinterwordspacing

\bibitem{chen_tun_chou_multi-robot_2011}
\BIBentryALTinterwordspacing
{Chen Tun Chou}, {Jiun-Yi Li}, {Ming-Fang Chang}, and {Li Chen Fu},
  ``Multi-robot cooperation based human tracking system using laser range
  finder,'' in \emph{2011 {IEEE} International Conference on Robotics and
  Automation}.\hskip 1em plus 0.5em minus 0.4em\relax {IEEE}, pp. 532--537.
  [Online]. Available: \url{http://ieeexplore.ieee.org/document/5980484/}
\BIBentrySTDinterwordspacing

\bibitem{ong_sensor_2012}
\BIBentryALTinterwordspacing
K.~S. Ong, Y.~H. Hsu, and L.~C. Fu, ``\BIBforeignlanguage{en}{Sensor fusion
  based human detection and tracking system for human-robot interaction},'' in
  \emph{\BIBforeignlanguage{en}{2012 {IEEE}/{RSJ} {International} {Conference}
  on {Intelligent} {Robots} and {Systems}}}.\hskip 1em plus 0.5em minus
  0.4em\relax Vilamoura-Algarve, Portugal: IEEE, Oct. 2012, pp. 4835--4840.
  [Online]. Available: \url{http://ieeexplore.ieee.org/document/6386222/}
\BIBentrySTDinterwordspacing

\bibitem{zhang_optimal_2016}
\BIBentryALTinterwordspacing
L.~Zhang and R.~Vaughan, ``Optimal robot selection by gaze direction in
  multi-human multi-robot interaction,'' in \emph{2016 {IEEE}/{RSJ}
  International Conference on Intelligent Robots and Systems ({IROS})}.\hskip
  1em plus 0.5em minus 0.4em\relax {IEEE}, pp. 5077--5083. [Online]. Available:
  \url{http://ieeexplore.ieee.org/document/7759745/}
\BIBentrySTDinterwordspacing

\bibitem{weinberg_creation_2009}
G.~Weinberg, B.~Blosser, T.~Mallikarjuna, and A.~Raman, ``The creation of a
  multi-human, multi-robot interactive jam session,'' p.~4.

\bibitem{iqbal_coordination_2017}
\BIBentryALTinterwordspacing
T.~Iqbal and L.~D. Riek, ``\BIBforeignlanguage{en}{Coordination {Dynamics} in
  {Multihuman} {Multirobot} {Teams}},'' \emph{\BIBforeignlanguage{en}{IEEE
  Robotics and Automation Letters}}, vol.~2, no.~3, pp. 1712--1717, Jul. 2017.
  [Online]. Available: \url{http://ieeexplore.ieee.org/document/7862739/}
\BIBentrySTDinterwordspacing

\bibitem{beer_framework_2017}
R.~D. Beer, C.~A. Rieth, R.~Tran, and M.~B. Cook,
  ``\BIBforeignlanguage{en}{Framework for {Multi}-{Human} {Multi}-{Robot}
  {Interaction}: {Impact} of {Operational} {Context} and {Team} {Configuration}
  on {Interaction} {Task} {Demands}},'' \emph{\BIBforeignlanguage{en}{2017 AAAI
  Spring Symposium Series}}, p.~8, 2017.

\bibitem{malik_developing_2019}
\BIBentryALTinterwordspacing
A.~A. Malik and A.~Bilberg, ``\BIBforeignlanguage{en}{Developing a reference
  model for human–robot interaction},''
  \emph{\BIBforeignlanguage{en}{International Journal on Interactive Design and
  Manufacturing (IJIDeM)}}, Jun. 2019. [Online]. Available:
  \url{http://link.springer.com/10.1007/s12008-019-00591-6}
\BIBentrySTDinterwordspacing

\bibitem{you_teaming_2017}
S.~You and L.~Robert, ``\BIBforeignlanguage{en}{Teaming up with {Robots}: {An}
  {IMOI} ({Inputs}-{Mediators}-{Outputs}-{Inputs}) {Framework} of
  {Human}-{Robot} {Teamwork}},'' p.~7, 2017.

\bibitem{tsarouchi_humanrobot_2017}
\BIBentryALTinterwordspacing
P.~Tsarouchi, G.~Michalos, S.~Makris, T.~Athanasatos, K.~Dimoulas, and
  G.~Chryssolouris, ``\BIBforeignlanguage{en}{On a human–robot workplace
  design and task allocation system},''
  \emph{\BIBforeignlanguage{en}{International Journal of Computer Integrated
  Manufacturing}}, vol.~30, no.~12, pp. 1272--1279, Dec. 2017. [Online].
  Available:
  \url{https://www.tandfonline.com/doi/full/10.1080/0951192X.2017.1307524}
\BIBentrySTDinterwordspacing

\bibitem{freedy_multiagent_2008}
\BIBentryALTinterwordspacing
A.~Freedy, O.~Sert, E.~Freedy, J.~McDonough, G.~Weltman, M.~Tambe, T.~Gupta,
  W.~Grayson, and P.~Cabrera, ``\BIBforeignlanguage{en}{Multiagent {Adjustable}
  {Autonomy} {Framework} ({MAAF}) for multi-robot, multi-human teams},'' in
  \emph{\BIBforeignlanguage{en}{2008 {International} {Symposium} on
  {Collaborative} {Technologies} and {Systems}}}.\hskip 1em plus 0.5em minus
  0.4em\relax Irvine, CA, USA: IEEE, May 2008, pp. 498--505. [Online].
  Available: \url{http://ieeexplore.ieee.org/document/4543970/}
\BIBentrySTDinterwordspacing

\bibitem{jones_dynamically_2006}
\BIBentryALTinterwordspacing
E.~Jones, B.~Browning, M.~Dias, B.~Argall, M.~Veloso, and A.~Stentz,
  ``\BIBforeignlanguage{en}{Dynamically formed heterogeneous robot teams
  performing tightly-coordinated tasks},'' in
  \emph{\BIBforeignlanguage{en}{Proceedings 2006 {IEEE} {International}
  {Conference} on {Robotics} and {Automation}, 2006. {ICRA} 2006.}}\hskip 1em
  plus 0.5em minus 0.4em\relax Orlando, FL, USA: IEEE, 2006, pp. 570--575.
  [Online]. Available: \url{http://ieeexplore.ieee.org/document/1641771/}
\BIBentrySTDinterwordspacing

\bibitem{Pinciroli:SI2012}
C.~Pinciroli, V.~Trianni, R.~O'Grady, G.~Pini, A.~Brutschy, M.~Brambilla,
  N.~Mathews, E.~Ferrante, G.~{Di Caro}, F.~Ducatelle, M.~Birattari, L.~M.
  Gambardella, and M.~Dorigo, ``{ARGoS}: a modular, parallel, multi-engine
  simulator for multi-robot systems,'' \emph{Swarm Intelligence}, vol.~6,
  no.~4, pp. 271--295, 2012.

\bibitem{vuforia}
``{Vuforia} augmented reality,'' \url{http://vuforia.com}, accessed:.

\bibitem{engine2008unity}
U.~G. Engine, ``Unity game engine-official site,'' \emph{Online][Cited: October
  9, 2008.] http://unity3d. com}, pp. 1534--4320, 2008.

\bibitem{hart_development_1988}
\BIBentryALTinterwordspacing
S.~G. Hart and L.~E. Staveland, ``\BIBforeignlanguage{en}{Development of
  {NASA}-{TLX} ({Task} {Load} {Index}): {Results} of {Empirical} and
  {Theoretical} {Research}},'' in \emph{\BIBforeignlanguage{en}{Advances in
  {Psychology}}}.\hskip 1em plus 0.5em minus 0.4em\relax Elsevier, 1988,
  vol.~52, pp. 139--183. [Online]. Available:
  \url{http://linkinghub.elsevier.com/retrieve/pii/S0166411508623869}
\BIBentrySTDinterwordspacing

\bibitem{black1976partial}
D.~Black, ``Partial justification of the borda count,'' \emph{Public Choice},
  pp. 1--15, 1976.

\bibitem{selcon1991workload}
S.~Selcon, R.~Taylor, and E.~Koritsas, ``Workload or situational awareness?:
  Tlx vs. sart for aerospace systems design evaluation,'' in \emph{Proceedings
  of the Human Factors Society Annual Meeting}, vol.~35, no.~2.\hskip 1em plus
  0.5em minus 0.4em\relax SAGE Publications Sage CA: Los Angeles, CA, 1991, pp.
  62--66.

\bibitem{schaefer2016measuring}
K.~E. Schaefer, ``Measuring trust in human robot interactions: Development of
  the “trust perception scale-hri”,'' in \emph{Robust Intelligence and
  Trust in Autonomous Systems}.\hskip 1em plus 0.5em minus 0.4em\relax
  Springer, 2016, pp. 191--218.

\bibitem{friedman1937use}
M.~Friedman, ``The use of ranks to avoid the assumption of normality implicit
  in the analysis of variance,'' \emph{Journal of the american statistical
  association}, vol.~32, no. 200, pp. 675--701, 1937.

\bibitem{roundtree2019transparency}
K.~A. Roundtree, M.~A. Goodrich, and J.~A. Adams, ``Transparency: Transitioning
  from human--machine systems to human-swarm systems,'' \emph{Journal of
  Cognitive Engineering and Decision Making}, p. 1555343419842776, 2019.

\end{thebibliography}

\end{document}